%% file: main.tex
\documentclass[10pt,twocolumn,letterpaper]{article}

\usepackage{cvpr}              % To produce the CAMERA-READY version
\usepackage{graphicx}
\usepackage{amsmath}
\usepackage{amssymb}
\usepackage{booktabs}
\usepackage{multirow}

\usepackage[pagebackref,breaklinks,colorlinks]{hyperref}

\usepackage[capitalize]{cleveref}
\crefname{section}{Sec.}{Secs.}
\Crefname{section}{Section}{Sections}
\Crefname{table}{Table}{Tables}
\crefname{table}{Tab.}{Tabs.}

\def\cvprPaperID{7851} % *** Enter the CVPR Paper ID here
\def\confName{CVPR}
\def\confYear{2023}

\begin{document}

% \title{Cameras Beating LiDAR 3D detection with Collaboration}
% \title{Collaboration Is What You Need for Camera to Overtake LiDAR in 3D Detection}
\title{Collaboration Helps Camera Overtake LiDAR in 3D Detection}
% \and
\author{Yue~Hu$^1$ \quad Yifan~Lu$^{1}$ \quad Runsheng~Xu$^2$ \quad Weidi~Xie$^{1,3}$ \quad Siheng Chen$^{1,3}$\footnotemark[1] \quad Yanfeng Wang$^{3,1}$ \\
$^{1}${Cooperative Medianet Innovation Center, Shanghai Jiao Tong University} \\ \quad $^{2}${University of California, Los Angeles} \quad $^{3}${Shanghai AI Laboratory} \\ 
$^{1}${\tt\small \{18671129361, yifan\_lu, weidi, sihengc, wangyanfeng\}@sjtu.edu.cn}  \quad 
$^{2}${\tt\small \{rxx3386\}@ucla.edu} \\
}

% \author{%
%   Yue~Hu^1, Yifan Lu^1, Runsheng Xu^2, {Weidi Xie}^1, {Siheng ~Chen}^1\thanks{Corresponding author}, {Yanfeng Wang}^1 \\
%   ^{1} Cooperative Medianet Innovation Center, Shanghai Jiao Tong University, Shanghai AI Laboratory \\
%    ^{2} University of California, Los Angeles \\
%    \texttt{18671129361, yifan\_lu,weidi, sihengc, wangyanfeng}@sjtu.edu.cn \\
%   \texttt{rxx3386}@ucla.edu \\
% }

% \author{%
%   Yue~Hu \hspace{1.75cm} Yifan Lu \\
%   Cooperative Medianet Innovation Center, Shanghai Jiao Tong University \\
%   \texttt{\{18671129361, yifan\_lu}@sjtu.edu.cn \\
%   Runsheng Xu University of California, Los Angeles 
%   \texttt{rxx3386@ucla.edu} \\
%   \and 
%    Weidi Xie, Siheng ~Chen\thanks{Corresponding author}, Yanfeng Wang\\
%   Shanghai Jiao Tong University,  Shanghai AI Laboratory \\
%   \texttt{weidi, sihengc, wangyanfeng}@sjtu.edu.cn
% }
\maketitle
\renewcommand{\thefootnote}{\fnsymbol{footnote}}
\footnotetext[1]{Corresponding author.}

\input{contents/0-abstract.tex}

\input{contents/1-introduction.tex}
\input{contents/2-related_work.tex}
\input{contents/3-method.tex}
\input{contents/4-experiments.tex}
\input{contents/5-conclusion.tex}

%%%%%%%%% REFERENCES
{\small
\bibliographystyle{ieee_fullname}
\bibliography{egbib}
}

\clearpage
\input{contents/6-supp.tex}

\end{document}

%% file: contents/0-abstract.tex
\begin{abstract}

Camera-only 3D detection provides an economical solution with a simple configuration for localizing objects in 3D space compared to LiDAR-based detection systems. However, a major challenge lies in precise depth estimation due to the lack of direct 3D measurements in the input. 
Many previous methods attempt to improve depth estimation through network designs, e.g., deformable layers and larger receptive fields.
This work proposes an orthogonal direction, improving the camera-only 3D detection by introducing multi-agent collaborations. 
Our proposed collaborative camera-only 3D detection (\texttt{CoCa3D}) enables agents to share complementary information with each other through communication.
Meanwhile, we optimize communication efficiency by selecting the most informative cues.
The shared messages from multiple viewpoints disambiguate the single-agent estimated depth and complement the occluded and long-range regions in the single-agent view.
We evaluate~\texttt{CoCa3D} in one real-world dataset and two new simulation datasets. Results show that~\texttt{CoCa3D} improves previous SOTA performances by 44.21\% on DAIR-V2X, 30.60\% on OPV2V+, 12.59\% on CoPerception-UAVs+ for AP@70.
Our preliminary results show a potential that with sufficient collaboration, the camera might overtake LiDAR in some practical scenarios. We released the \href{https://siheng-chen.github.io/dataset/CoPerception+}{dataset} and \href{https://github.com/MediaBrain-SJTU/CoCa3D}{code}.
\end{abstract}

%% file: contents/1-introduction.tex
\vspace{-3mm}
\section{Introduction}
\vspace{-1mm}

As a fundamental task of computer vision, 3D object detection aims to localize objects in the 3D physical space given an agent's real-time sensor inputs. It is crucial in a wide range of applications, including autonomous driving~\cite{lang2019pointpillars,zhou2022cross,Wu2020MotionNetJP}, surveillance systems~\cite{Yin2021Centerbased3O}, robotics~\cite{Li2022Dualview3O} and unmanned aerial vehicles~\cite{Hu2022AM3D}. Depending on sensor setups, there are multiple technical solutions to realize 3D object detection. 
In this spectrum, one extreme emphasizes raising the upper bound of detection performance, which uses high-end LiDAR sensor~\cite{zhou2018voxelnet,lang2019pointpillars,shi2020pv,deng2021voxel,zheng2021cia,mao2021pyramid,Chen20213DPC} to collect precise 3D measurements. However, this approach is too expensive to scale up. The other extreme solution emphasizes cost effectiveness, which tries to use thrifty sensor setups, e.g. only using cameras to detect 3D objects in real-time~\cite{wang2022detr3d,philion2020lift, zhang2022beverse,huang2021bevdet, wang2022dfm,roddick2018orthographic,carion2020end,zhou2022cross,huang2022bevdet4d,CaDDN}. 
However, camera-only 3D detection is significantly and consistently worse than LiDAR-based detection in most scenarios~\cite{Liu2022BEVFusionMM}.

\begin{figure}
\vspace{-2mm}
    \centering
    \includegraphics[width=0.99\linewidth]{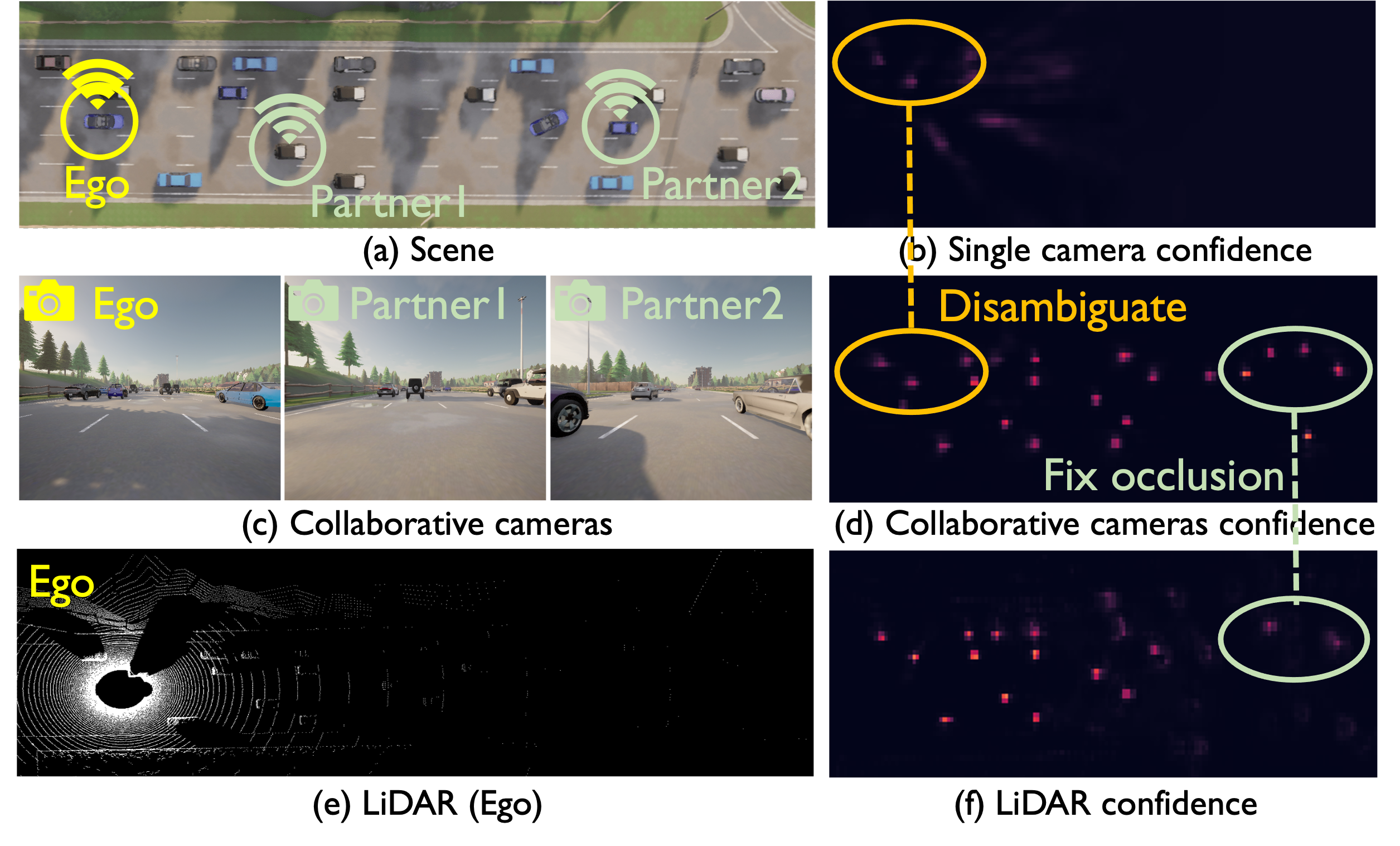}
  \vspace{-4mm}
  \caption{Collaborative camera-only 3D detection can disambiguate the single-view estimated depth, address the long-range and occlusion issues, and approach LiDAR in 3D detection.}
  \vspace{-6mm}
  \label{fig:intro}
\end{figure}

In this paper, we propose an orthogonal direction for improving camera-only 3D detection performances by introducing multi-agent collaborations. 
Hypothetically, empowered by advanced communication systems, multiple agents equipped only with cameras could share visual information with each other. 
% \weidi{would it be helpful to give some high-level intuition on why multi-agent can help, objects that are far away or invisible from one agent might be evident for another agent at other spatial location, ....}
This would bring three outstanding benefits. 
First, different viewpoints from multiple agents can largely resolve the depth ambiguity issue in camera-only 3D detection, bridging the gap with expensive LiDARs on depth estimation.
Second, multi-agent collaboration avoids inevitable limitations in single-agent 3D detection, such as occlusion and long-range issues, and potentially enables more holistic 3D detection; that is, detecting all the objects existed in the 3D scene, including those beyond visual range. Since LiDAR also suffers from limited field of view, this potentially enables collaborative cameras to outperform LiDAR. Third, the total expense of a large fleet of vehicles is significantly reduced as cameras are much cheaper than LiDAR.  However, multi-agent collaboration also brings new challenges. Different from many multi-view geometry problems, here we also have to concern communication bandwidth constraints. Thus, each agent needs to select the most informative cues to share.

Following this design rationale, 
we propose a novel collaborative camera-only 3D detection framework~\texttt{CoCa3D}. It includes three parts:
i) single-agent camera-only 3D detection, which achieves basic depth estimation and 3D detection for each agent; 
ii) collaborative depth estimation, which disambiguates the estimated depths by promoting spatial consistency across multiple agents’ viewpoints; and iii) collaborative detection feature learning, which complements detection features by sharing key detection messages with each other.
Compared to recent collaborative perception methods~\cite{li2021disconet,lei2022SyncNet} that are dealing with LiDAR,~\texttt{CoCa3D} specifically designs novel collaborative depth estimation to customize the task of camera-only 3D detection.

To evaluate~\texttt{CoCa3D}, we conduct comprehensive experiments on one real-world dataset, DAIR-V2X~\cite{yu2022dairv2x}, and two new simulation datasets, OPV2V+ and CoPerception-UAVs+, which are extended based on original OPV2V~\cite{xu2022opv2v} and CoPerception-UAVs~\cite{hu2022where2comm} with more collaborative agents that cover three types of agents (cars, infrastructures and drones). 
Our results show that i) with 10 collaborative agents, \texttt{CoCa3D} enables camera-only detectors to overtake LiDAR-based detectors on OPV2V+; and ii) \texttt{CoCa3D} consistently outperforms previous works in the performance-bandwidth trade-off across multiple datasets by a large margin, improving the previous SOTA performances by 30.60\% on OPV2V+, 12.59\% on CoPerception-UAVs+, 44.21\% on DAIR-V2X for AP@70.
To sum up, our contributions are:

$\bullet$ We propose a novel collaborative camera-only 3D detection framework~\texttt{CoCa3D}, which improves the detection ability of cameras with multi-agent collaboration, promoting more holistic 3D detection.

$\bullet$ We propose core communication-efficient collaboration techniques, which explore the spatially sparse yet critical depth messages and tackle the depth ambiguity, occlusion, and long-range issues by fusing complementary information from different viewpoints, achieving more accurate and complete 3D representation.
% i) collaborative depth estimation, which improves depth estimation by leveraging multi-view consistency and reduces the communication cost by focusing on critical areas, ii) collaborative detection feature learning, which improves detection features by leveraging multi-view complementary and reduces the communication cost by focusing on critical areas.

$\bullet$ We expand two previous collaborative datasets with more agents, and conduct extensive experiments, validating that i)~\texttt{CoCa3D} significantly bridges the performance gap between camera and LiDAR on OPV2V+ and DAIR-V2X; and ii)~\texttt{CoCa3D} achieves the state-of-the-art performance-bandwidth trade-off.

%% file: contents/2-related_work.tex
\vspace{-3mm}
\section{Related Work}
\vspace{-1mm}

\noindent\textbf{Camera-only 3D object detection.}
Camera-only 3D detection aims to detect objects in the 3D space given the 2D image through explicit or implicit depth estimation~\cite{roddick2018orthographic, philion2020lift, zhang2022beverse,huang2021bevdet, wang2022dfm,Hu2022AM3D}. Recently, bird’s-eye-view (BEV) representations are widely used for their computation efficiency and comparable performance compared to the 3D voxel features~\cite{lang2019pointpillars}. 
To get the BEV features, there are two types of methods. The depth-based methods~\cite{philion2020lift,huang2022bevdet4d, CaDDN} first estimate the depth distribution, and then attentively project the 2D image features along the projection ray to get the 3D voxel features and collapse it to BEV feature. The query-based methods~\cite{peng2022bevsegformer, li2022bevformer,zhou2022cross} first initialize queries for each BEV grid and then leverage transformer-architecture-based cross-attention to query image features with camera-aware positional embedding. Here, our single-agent camera-only detector follows the simple-yet-effective CaDDN~\cite{CaDDN}.

\noindent\textbf{LiDAR 3D object detection.}
LiDAR-based 3D detection achieves excellent performance due to the precise 3D measurements of the input data. Two well-known ways to encode LiDAR points include voxel-based~\cite{zhou2018voxelnet,lang2019pointpillars,zheng2021cia} and point-based~\cite{shi2020pv, shi2021pv,yang20203dssd}. Voxel-based methods divide the 3D space into regular voxels~\cite{zhou2018voxelnet} or pillars~\cite{lang2019pointpillars}, and encode the point inside into feature representations. Point-based methods are usually based on the PointNet~\cite{qi2017pointnet} series to aggregate the feature of points. Then the point features will be used for the proposal refinement. 
LiDAR-based 3D detection performs well, but high-quality LiDAR is hard to be adopted in a large scale due to the expensive cost. Here we propose an economic camera-only solution by introducing multi-agent collaboration, whose performance can catch up with LiDAR given a sufficient number of agents.
% Voxel-based representations are easier to parallelize, which motivates our use. Here, we project the 2D image features given the estimated depth into 3D voxels and take advantage of the voxel representation learning skills.

\noindent\textbf{Collaborative perception.}
Collaborative perception is an emerging application of multi-agent systems, which promotes perception performance by enabling agents to share information with other agents through communication. In this emerging research field, a surge of high-quality datasets, (V2X-SIM~\cite{Li_2021_RAL}, OPV2V~\cite{xu2022opv2v}, DAIR-V2X~\cite{yu2022dairv2x}, CoPerception-UAVs~\cite{hu2022where2comm}), and collaborative methods~\cite{who2com,liu2020when2com,wang2020v2vnet,xu2022opv2v,li2021disconet,Li_2021_RAL,xu2022v2xvit,yu2022dairv2x,xu2022CoBEVT,hu2022where2comm,lu2022robust,Li2022MultiRobotSC,Gao2020RegularizedGM} aimed for better performance-bandwidth trade-off have been proposed. 
Collaborative methods have three types: early, late and intermediate fusion. Compared with early fusion(i.e. transmitting raw data) and late fusion (i.e. transmitting outputs), intermediate fusion(i.e. transmitting and fusing features encoded by deep networks) achieves a better performance-bandwidth trade-off. DiscoNet~\cite{li2021disconet} adopts knowledge distillation to take the advantage of both early and intermediate collaboration. 
% V2X-ViT ~\cite{xu2022v2xvit} introduces a novel heterogeneous multi-agent attention module to fuse information across vehicles and infrastructures. 
Where2comm~\cite{hu2022where2comm} optimizes communication efficiency by sharing spatially sparse yet perceptually critical information. However, previous works mainly focus on LiDAR-based 3D object detection, while we investigate collaborative camera-only detection and specifically enhance depth estimation via multi-agent collaboration.

% LiDAR-based 3D object detection reaches high accuracy, but not all cars can be equipped with lidar due to its high price.
% It mainly has two types. The single-stage detectors~\cite{zhou2018voxelnet, lang2019pointpillars, zheng2021cia} have faster inference speed owing to the simple pipeline. The two-stage detectors~\cite{shi2020pv,deng2021voxel,mao2021pyramid} have higher accuracy owing to the repeated feature refinement. Both types tend to have voxel representation 

% VoxelNet~\cite{zhou2018voxelnet} divides point cloud into regular voxels, and transforms points within voxel into feature representation. PointPillars ~\cite{lang2019pointpillars} encodes point cloud into standing pillars and apply 2D convolution on BEV view feature map. CIA-SSD\cite{zheng2021cia} designs novel SSFA module and IoU-aware confidence rectification module to improve detection accuracy. For two-stage model, carefully designed features  will be used for box refinement. PV-RCNN~\cite{shi2020pv} utilizes voxel set abstraction module to encode the scene by keypoints, and refine proposals by RoI-grid pooling. Voxel-RCNN~\cite{deng2021voxel} proposes voxel RoI pooling for aggregating features within proposals. \Note{sc: relation to us?}

%% file: contents/3-method.tex
\begin{figure*}
    \centering
    \includegraphics[width=0.99\linewidth]{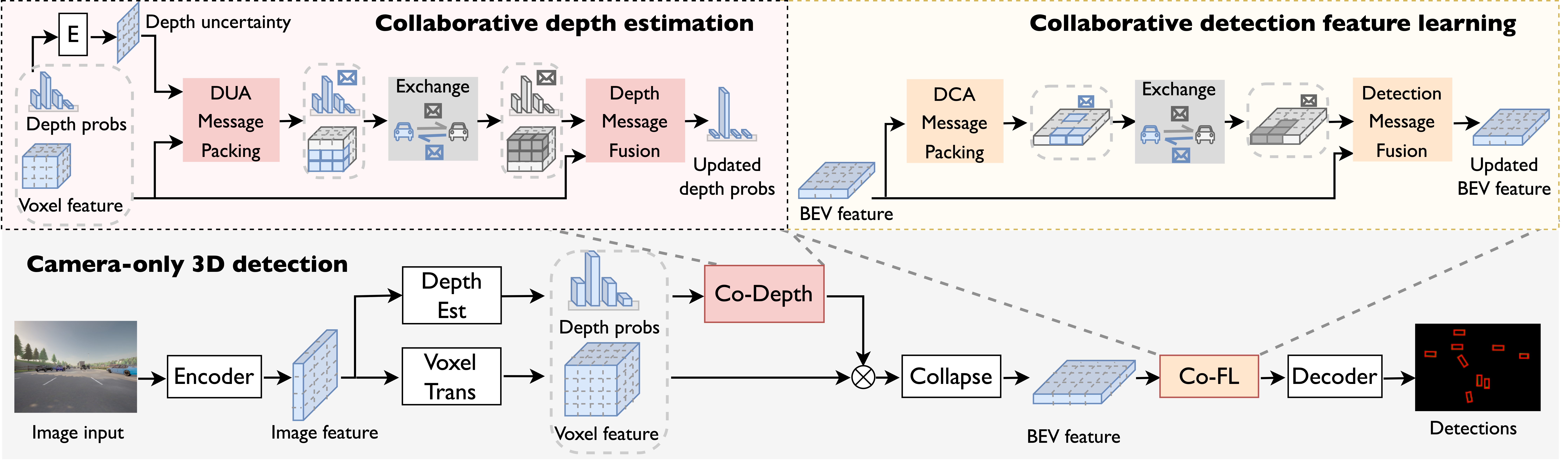}
  \vspace{-3mm}
  \caption{System overview. \texttt{CoCa3D} is a camera-only 3D detector integrated with two collaboration modules. Collaborative depth estimation (Co-Depth) enhances the single-agent estimated depth to achieve more accurate 3D feature. Collaborative detection feature learning (Co-FL) complements the single-agent 3D feature to achieve more holistic 3D detection.}
  \vspace{-5mm}
  \label{fig:framework}
\end{figure*}

\vspace{-2mm}
\section{Collaborative Camera-Only 3D Detection}
\vspace{-1mm}

This section presents~\texttt{CoCa3D}, 
collaborative camera-only 3D detection framework, 
which enables multiple agents to share visual information with each other, promoting more holistic 3D detection.

\subsection{Problem Formulation}
\vspace{-1mm}
Consider $N$ agents in the scene, 
let $\mathcal{X}_i$ be the RGB image collected by the $i$th agent and $\mathcal{O}^0_i$ be the corresponding ground-truth detection. The objective of~\texttt{CoCa3D} is to maximize detection performances of all agents given certain communication budget $B$; that is,
\setlength{\abovedisplayskip}{2pt}
\setlength{\belowdisplayskip}{2pt}
\begin{small}
\begin{eqnarray*}
    \underset{\theta,\mathcal{P}}{\max}~\sum_{i} 
     g \left(\Phi_{\theta} \left(\mathcal{X}_i,\{\mathcal{P}_{i\rightarrow j} \}_{j=1}^N \right), \mathcal{O}^0_i  \right),
    {\rm s.t.~} \sum_{i} |\mathcal{P}_{i\rightarrow j} | \leq B,
\end{eqnarray*}
\end{small}
where $g(\cdot,\cdot)$ is the detection evaluation metric, $\Phi(\cdot)$ is a detection model with trainable parameter $\theta$, and $\mathcal{P}_{i\rightarrow j}$ is the message transmitted from the $i$th agent to the $j$th agent. The key challenge is to determine the messages $\mathcal{P}_{i\rightarrow j}$, which should be both informative and compact. 
Our design rational comes from two aspects: First, since the major gap between camera and LiDAR is the depth, the message should include depth information. This would allow different viewpoints from multiple agents to disambiguate the infinite depth possibilities and localize the correct depth candidate. Second, the message should include detection clues to provide complementary detection information, which can fundamentally overcome inevitable limitations of single-agent detection, such as occlusion and long-range issues.

% The motivation is that single-agent camera-only 3D object detection is an ill-posed task as each image pixel corresponds to a projected ray in the physical world, that is, there are infinite depth possibilities in the 3D space. Furthermore, single-agent encounters inevitable physical limitations, such as limited field of view, occlusion, and long-range issues. Multi-agent collaboration can fundamentally address these physical limitations: a) different viewpoints from multiple agents can disambiguate the infinite depth possibilities and localize the correct depth candidate; b) multi-view compensation enables more accurate and holistic 3D representations.

Based on the above intuition, the proposed~\texttt{CoCa3D} includes a) single-agent camera-only 3D detection, which achieves basic  depth estimation and detection ability, and b) multi-agent collaboration, which shares both estimated depth information and detection features to improve 3D representation and detection performances; see Fig.~\ref{fig:framework}.

% \texttt{CoCa3D} includes three parts: a) single-agent camera-only 3D object detection, which follows the monocular 3D object detector~\cite{CaDDN}. It learns to project image features into 3D space by estimating depth distribution;
% b) collaborative depth estimation, which improves the single-agent depth estimation with multi-view geometry to generate more accurate 3D feature;
% and c) collaborative detection feature learning, which improves the single-agent 3D feature with the shared compact complementary messages.

\vspace{-1mm}
\subsection{Single-agent camera-only 3D detection}
\vspace{-1mm}
Single-agent camera-only 3D object detection network learns to detect 3D objects in the physical space based on the 2D camera inputs. Here we adopt CaDNN~\cite{CaDDN}, an off-the-shelf architecture. The main idea is to project the flat image feature into all the possible depths in the 3D space, and then attentively aggregate the 3D voxel feature and collapse them into the BEV feature. To mention that, the BEV feature is used because it largely reduces the computation cost, while performing similarly to 3D voxel features~\cite{lang2019pointpillars}. 
The architecture consists of five modules as detailed below.

 % {\HC  The \textbf{encoder} extracts features from the observed images. Based on the image features, the \textbf{depth estimation} module estimates the depth distribution and the uncertainty for each spatial region and the \textbf{voxel transformation} module lifts the flat image feature to the 3D voxel space via the camera calibration parameters. Then the voxel features are weighted based on the depth distribution and collapsed to BEV feature in the \textbf{collapse} module. The \textbf{decoder} decodes the BEV feature into object detections.}\Note{SC: not your contribution, too long-winded}

\vspace{2pt}
\noindent \textbf{Encoder.} The encoder extracts features from the input images. For agent $i$, given its input $\mathcal{X}_i$, the output feature map of encoder is 
$ \mathcal{F}_i = \Phi_{\rm enc}(\mathcal{X}_i) \in \mathbb{R}^{H \times W \times C},$ where $H,W,C$ are the height, width and channel of the image feature.

\vspace{2pt}
\noindent \textbf{Depth estimation.} The depth estimation module predicts pixel-wise categorical depth distribution over a set of predefined depth bins to accurately locate image information in 3D space. Treating the estimation as a classification problem can capture the inherent depth estimation uncertainty to reduce the impact of erroneous depth estimates, as stated in CaDNN~\cite{CaDDN}.
To achieve this, a parametric depth estimation network $\Phi_{\rm Depth}(\cdot)$ is used, and the pixel-wise categorical depth distribution is obtained as
% \vspace{-2mm}
\begin{equation}
\label{eq:depth}
    \mathcal{D}_i =\Phi_{\rm Depth}(\mathcal{F}_i)\in\mathbb{R}^{H \times W \times D},    
\end{equation}
whose $(h,w)$th element $\mathcal{D}_i(h,w) \in \mathbb{R}^{D}$ refers to the estimated depth distribution of the $(h,w)$th image pixel with $D$ as the number of discretized depth bins.

\vspace{2pt}
\noindent \textbf{Voxel transformation.} Voxel transformation projects the image feature to the 3D space, given all the possible depth candidates and the known image calibration matrix, resulting in a 3D voxel representation $\widehat{\mathcal{V}}_i\in\mathbb{R}^{X\times Y\times Z \times C}$. The camera projection matrix $\mathbf{P}\in\mathbb{R}^{3\times 4}$ defines the mapping between the 3D voxel coordinate to the image pixel coordinate, this is, $[u,v,d]=\mathbf{P}\cdot[x,y,z,1]$. Similarly, the depth probabilities can be transformed to the 3D voxel space, resulting in $\widehat{\mathcal{D}}_i\in\mathbb{R}^{X \times Y \times Z}$, where each element indicates the confidence that the feature pixel belongs to the voxel.

\vspace{2pt}
\noindent \textbf{Collapse.} The voxel feature is collapsed to a single height plane to generate the bird's-eye-view (BEV) feature as $\overline{\mathcal{B}}_i=\Phi_{\rm clp}( \widehat{\mathcal{D}}_i \odot \widehat{\mathcal{V}}_i)\in\mathbb{R}^{X\times Y \times C}$, where $\odot$ is the element-wise multiplication. Symbol $\cdot$ / $\widehat{\cdot}$ / $\overline{\cdot}$ reflect 2D image / 3D voxel / BEV spaces, respectively. The collapse network $\Phi_{\rm clp}(\cdot)$ flattens the 3D voxel feature along the $Z$-axis and then applies $1\times 1$ convolution to reduce the channel dimension.

\noindent \textbf{Decoder.} 
The decoder takes the BEV feature as input, and outputs the objects including class and regression. Here, we adopt CenterNet~\cite{zhou2019objects}, an off-the-shelf detector. 
Given the BEV feature $\overline{\mathcal{B}}_i$, the detection decoder $\Phi_{\rm dec}(\cdot)$ generate the dense heatmap of $i$th agent by
$
     \overline{\mathcal{O}}_i  =  \Phi_{\rm dec}(\overline{\mathcal{B}}_i) \in \mathbb{R}^{X \times Y \times 7},
$
where each location of $\overline{\mathcal{O}}_i$ represents a rotated box with class $(c,x,y,h,w, \cos\alpha, \sin\alpha)$, denoting class confidence, position residual, size and angle. Non-maximum suppression (NMS) is applied to the dense predictions and generate the final sparse output of the 3D detection system. 

\vspace{2pt}
\noindent \textbf{Discussion.}
% \weidi{add a small discussion section, on what is the problems of the single-agent 3D detection, and build a link for your next section, say, which of this pipeline our collaborative approach can improve significantly, for example, depth estimation~($\hat{\mathcal{D}}$)......}
Admittedly, single-agent camera-only 3D detection makes localizing objects in the 3D space given 2D image input possible through numerous ingenious algorithm designs. The 3D detection performance is still far from perfect due to the ambiguous single-agent depth estimation and limited visual range. Tackling these two issues are challenging for a single agent, but from a perspective of multi-agent collaboration, sharing complementary visual clues makes depth estimation and visual range enlargement much more natural. This motivates our following designs.

\vspace{-1mm}
\subsection{Multi-agent collaboration} 
\vspace{-1mm}
% The proposed multi-agent collaboration module targets to improve the single-agent 3D representation and achieve more accurate and holistic 3D detection through information sharing between agents. 
Multi-agent collaboration includes two parts: i) collaborative depth estimation, which enables the sharing of depth information estimated by individual agents to produce more accurate 3D representation and less aliased BEV feature; and ii) collaborative detection feature learning, which enables the sharing of detection features of individual agents to achieve more holistic 3D detection.  

\vspace{-4mm}
\subsubsection{Collaborative depth estimation}
\vspace{-1mm}
Collaborative depth estimation (\textbf{Co-Depth}) targets to disambiguate infinite depth possibilities in single-agent camera-only depth estimation and localize the correct depth candidate through multi-view consistency.
The intuition is, for a correct depth candidate, its corresponding 3D location should be spatially consistent from multiple agents' viewpoints.
To achieve this, each agent can exchange the depth information through communication. Meanwhile, we promote communication efficiency by selecting the most critical and unambiguous depth information.
% , and take advantage of multi-view consistency to refine the single-agent estimated depth.
% However, communication systems in real-world scenarios are always constrained and they can hardly afford huge communication consumption in real time. 
% Therefore, we should exchange the most critical and unambiguous depth information to make the best use of the precious communication bandwidth. 
Accordingly, Co-Depth includes: a) depth uncertainty aware message packing, which packs compact messages with unambiguous depth information; and b) depth message fusion, which enhances depth estimation with the received depth messages.

\vspace{2pt}
\noindent \textbf{Depth uncertainty aware message packing.}
Depth uncertainty-aware (DUA) message packing packs the most critical depth information used for multi-view consistency into the to-be-sent message based on the depth uncertainty.  
The depth message includes: i) the voxel feature, which is used for the multi-view visual similarity measurement; and ii) the depth probability, which indicates the confidence that the feature pixel belongs to the voxel and is used for the multi-view candidate selection.

The spatially sparse features of $i$th agent are selected based on its spatial depth uncertainty map in the 2D image space $\mathbf{U}_{i}\in \mathbb{R}^{H\times W}$, that can be represented by the pixel-wise entropy from corresponding depth distribution,
%where each element at the image pixel $(h,w)$ is obtained through the entropy of the corresponding depth distribution, this is, 
\setlength{\abovedisplayskip}{3pt}
\setlength{\belowdisplayskip}{3pt}
\begin{equation*}
    \mathbf{U}_{i}(h,w)={\rm H}(\mathcal{D}_i(h,w)) \in \mathbb{R},
\end{equation*}
where ${\rm H}(\cdot)$ is the entropy function and $\mathcal{D}_i(h,w)$ is the depth distribution at the $(h,w)$th image pixel, following from~\eqref{eq:depth}. The intuition is, for the locations with low uncertainty score, an agent is confident about which depth bin the feature pixel belongs to. Thus, sharing information at these locations can help with depth estimation accuracy. On the contrary, for the locations with high uncertainty score, an agent is hard to tell which depth bin the feature pixel belongs to. Transmitting these spatial features would introduce ambiguity, and even degrade the single-agent depth estimation.

Specifically, a binary selection matrix is used to represent whether each location is selected or not, where $1$ denotes selected, and $0$ elsewhere. Let $\mathbf{M}^d_{i\rightarrow j} \in \{0, 1\}^{H\times W}$ be a selection matrix to determine the depth message sent from agent $i$  to $j$. Its $(h,w)$th element is
\begin{equation*}
\label{eq:selector}
\mathbf{M}^d_{i\rightarrow j}(h,w)= I(\mathbf{U}_{i} (h,w)<u_{\rm thre}),
\end{equation*}
where $I(\cdot)$ is an indicator function and $u_{\rm thre}$ is a predefined threshold. This reflects that a specific spatial area will be selected when its depth uncertainty is below a threshold.

The selection matrix can be projected in the 3D voxel space, resulting in $\widehat{\mathbf{M}}^d_{i\rightarrow j}\in\mathbb{R}^{X\times Y \times Z}$.  Then, the selected voxel feature map $\widehat{\mathcal{Z}}_{i\rightarrow j}$ and  the selected depth distribution $\widehat{\mathcal{E}}_{i\rightarrow j}$ sent from the $i$th agent to the $j$th agent are obtained as
\begin{align*}
\widehat{\mathcal{Z}}_{i\rightarrow j}&=\widehat{\mathbf{M}}^d_{i\rightarrow j}\odot \widehat{\mathcal{V}}_i \in \mathbb{R}^{X\times Y \times Z \times C},\\
\widehat{\mathcal{E}}_{i\rightarrow j}&=\widehat{\mathbf{M}}^d_{i\rightarrow j}\odot \widehat{\mathcal{D}}_i \in \mathbb{R}^{X\times Y \times Z},    
\end{align*}
where $\widehat{\mathcal{V}}_i$ and $\widehat{\mathcal{D}}_i$ are the full 3D voxel feature map and depth distribution. Overall, the depth message sent from the $i$th agent to the $j$th agent is $\mathcal{P}^d_{i\rightarrow j}=(\widehat{\mathcal{E}}_{i\rightarrow j}, \widehat{\mathcal{Z}}_{i\rightarrow j})$, where both entries are spatially sparse and communication efficient.

% Note that i) $\mathcal{E}_{i\rightarrow j}$ provides spatial priors to request complementary information for the $i$th agent's need in the next round; the feature map $\mathcal{Z}_{i\rightarrow j}$ provides supportive information for the $i$th agent's need in the this round. They together enable mutually beneficial collaboration; ii) since  $\mathcal{Z}_{i\rightarrow j}^{(k)}$ is sparse, we only transmit non-zero features and corresponding indices, leading to low communication cost; and iii) the sparsity of $\mathcal{Z}_{i\rightarrow j}^{(k)}$ is determined by the binary selection matrix, which dynamically allocates the communication budget at various spatial areas based on their perceptual critical level, adapting to various communication conditions. 

\vspace{2pt}
\noindent \textbf{Depth message fusion.} Depth message fusion targets to enhance the depth estimation given the received depth messages through different  viewpoints of multiple agents. The intuition is that for a correct depth candidate, the visual features at the same 3D point observed by multiple agents should be similar. To achieve this, we introduce multi-view depth consistency weighting. Let $\widehat{\mathcal{S}}_i(x,y,z) \in \mathbb{R}$ be the matching score between agent $i$ and its neighbors $\mathcal{N}_i$, those who share messages with current agent, at coordinate $(x,y,z)$. The multi-view matching score is obtained as
\setlength{\abovedisplayskip}{1pt}
\setlength{\belowdisplayskip}{1pt}
\begin{small}
\begin{align*}
& \widehat{\mathcal{W}}_{j\rightarrow i}(x,y,z) = I(\widehat{\mathcal{E}}_{j\rightarrow i}(x,y,z)>p_{\rm thre})\in \{0, 1\}, \\
& \widehat{\mathcal{S}}_i(x,y,z) = \sum_{j\in \mathcal{N}_i} \widehat{\mathcal{W}}_{j\rightarrow i}(x,y,z)\left<\widehat{\mathcal{V}}_i(x,y,z), \widehat{\mathcal{Z}}_{j\rightarrow i}(x,y,z)\right>, 
\end{align*}    
\end{small}
where $\widehat{\mathcal{W}}_{j\rightarrow i}(x,y,z)$ is a binary weight used to filter out the neighboring view that has a low depth confidence, below a threshold $p_{\rm thre}$ and $\left<\cdot,\cdot\right>$ is the inner product that measures the visual feature consistency across two views. This means that either the depth candidate is wrong or it is not visible in that view (e.g. due to occlusion), where the multi-view matching score should not be considered. 
% Note that depth consistency weighting discards the candidates with low single-view depth probability. Such weighting is useful especially when the multi-view matching is ambiguous or unreliable. For example, if the pixel is within a texture-less surface, a wide range of depth candidates will lead to similar matching scores. If the scene contains reflective surfaces, the matching score will be computed between the reflections, resulting in over-estimated depth. In both cases, we can make robust prediction by favoring the depth candidates with high single-view depth probability.

Given the multi-view matching score, the collaborative depth distribution is obtained as
\begin{equation*}
    \widehat{\mathcal{D}}'_i = \Phi_{\rm CoDepth}\left(\left[\widehat{\mathcal{D}}_i; \widehat{\mathcal{S}}_i\right]\right)\in \mathbb{R}^{X \times Y \times Z},
\end{equation*}
where the $[;]$ denotes concatenation, $\Phi_{\rm CoDepth}(\cdot)$ is the collaborative depth estimation network implemented with $1\times1$ convolutions followed by a sigmoid activation. Each element of the collaborative depth distribution $\widehat{\mathcal{D}}'_i$ reflects the confidence of the feature locates in the corresponding voxel. Note that i) the collaboratively updated depth probability is decided on both the single-view confidence and the multi-view consistency; ii) the depth consistency weighting discards the candidates with low single-view depth probability to avoid ambiguity and make the estimation more robust.

Given this collaborative depth distribution, each agent can weigh the 3D voxel feature and then collapse into a BEV feature with a collapse network $\Phi_{\rm clp}(\cdot)$, this is, $\overline{B}_i=\Phi_{\rm clp}(\widehat{\mathcal{D}}'_i \odot \widehat{\mathcal{V}}_i)\in\mathbb{R}^{X\times Y \times C}$. The collapse network is implemented with $1\times 1$ convolution to reduce channel dimension.

\vspace{-4mm}
\subsubsection{Collaborative detection feature learning}
\vspace{-2mm}
Collaborative depth estimation carefully refines the depth and promotes more accurate 3D representations for each single agent. However, the physical limitations of single agent, such as, limited field of view, occlusion, and long-range issues still remain. To achieve more holistic 3D detection, each agent should be able to exchange 3D detection feature and make use of the complementary information. 
Meanwhile, we promote communication efficiency by selecting the most perceptually critical information.
Therefore, collaborative detection feature learning (\textbf{Co-FL}) includes: a) detection confidence aware message packing, which packs spatially sparse yet perceptually critical 3D features with the guidance of the detection confidence; and ii) detection message fusion, which enhances the 3D feature with the received detection messages.

\vspace{2pt}
\noindent \textbf{Detection confidence aware message packing.} The detection confidence aware (DCA) message packing targets to pack the complementary perceptual information into a compact message. The core idea is to explore the spatial heterogeneity of perceptual information. The intuition is that the areas that contain objects are more critical than background areas. During collaboration, areas with objects could help recover the miss-detected objects due to the limited view; and background areas could be omitted to save the precious bandwidth. To achieve this, we implement the spatial confidence map with the detection confidence map by
\begin{equation*}
    \overline{\mathbf{C}}_i=\Phi_{\rm cls}(\overline{\mathcal{B}}_i)\in\mathbb{R}^{X\times Y},
\end{equation*}
where $\Phi_{\rm cls}(\cdot)$ denotes the standard classification network in the single-agent decoder $\Phi_{\rm det}(\cdot)$, $\overline{\mathcal{B}}_i$ is the BEV feature generated with the collaboratively estimated depth. Based on this map, agents decide which spatial area to communicate; that is, each agent offers spatially sparse, yet critical features to support other agents. 

Specifically, a binary selection matrix is used to represent each location is selected or not. Let $\overline{\mathbf{M}}^f_{i\rightarrow j} \in \{0, 1\}^{X \times Y}$ be a selection matrix to determine the detection message sent from agent $i$ to $j$. Its $(x,y)$th element is
\begin{equation*}
\label{eq:selector}
\overline{\mathbf{M}}^f_{i\rightarrow j}(x,y)= I( \overline{\mathbf{C}}_i (x, y) > c_{\rm thre}),
\end{equation*}
where $I(\cdot)$ is an indicator function and $c_{\rm thre}$ is a predefined threshold. This reflects that a specific spatial area will be selected when its detection confidence is above a threshold.  Overall, the detection message sent from the $i$th agent to the $j$th agent is $\mathcal{P}^f_{i\rightarrow j}= \overline{\mathbf{M}}^f_{i\rightarrow j} \odot \overline{\mathcal{B}_i}$. Only selected features and their indices are packed, and they can be recovered at the receiver. This greatly reduces communication cost. 
% Agents exchange these compact detection messages with the partners.

\vspace{2pt}
\noindent \textbf{Detection message fusion.} Here we augment the detection feature of each agent by aggregating the received detection messages from other agents. We implement this with simple but effective non-parametric point-wise maximum fusion.
Specifically, for the $i$th agent, after receiving the $j$th agent's message $\mathcal{P}^f_{j\rightarrow i}$. We also include the ego feature map in fusion and denote $\mathcal{P}^f_{i\rightarrow i} = \overline{\mathcal{B}}_{i}$ to make the formulation simple and consistent, where $\mathcal{P}^f_{i\rightarrow i}$ might not be sparse. The fused BEV feature is obtained as
\begin{equation*}
\overline{\mathcal{B}}'_i=\underset{j\in\mathcal{N}_i\cup\{i\}}{\rm max}(\mathcal{P}^f_{j\rightarrow i})\in\mathbb{R}^{X\times Y \times C},
\end{equation*}
where ${\rm max}(\cdot)$ maximizes the corresponding features from multiple agents at each individual spatial location. Note that attention fusion is not permutation invariant, as attention weights vary with the ordering of key and query. Here we simply use the max operator to avoid this permutation variant issue. The fused BEV feature is output to the decoder to generate the final detection.
% \Note{sc: how to obtain the final detection after collaboration?}

% {\HC Note that we apply max fusion instead of attention fusion because with attention fusion, even with the same multi-view observation, the representation of the same region varies when choosing a different ego agent, which is not in line with the actual situation. ???} To avoid this issue, we apply the ego-invariant and more computation-efficient max fusion. 
\vspace{-2mm}
\subsection{Losses}
\vspace{-2mm}
To train the overall system, we supervise two tasks: categorical depth estimation and 3D object detection. The overall loss is
$
    L = L_{\rm dep} \left(\mathcal{D}_i,\mathcal{D}^0_i \right) + L_{\rm det} \left(\overline{\mathcal{O}}_i,\mathcal{O}^0_i \right),
$
where $\mathcal{D}^0_i$ and $\mathcal{O}^0_i$ is the $i$th agent's ground-truth depth category and objects, respectively, $L_{\rm dep}$ and $L_{\rm det}$ is the depth classification loss~\cite{CaDDN} and detection loss~\cite{zhou2019objects}, respectively.

%% file: contents/4-experiments.tex
\begin{figure*}[!t]
  \centering
    \begin{subfigure}{0.32\linewidth}
    \includegraphics[width=0.85\linewidth]{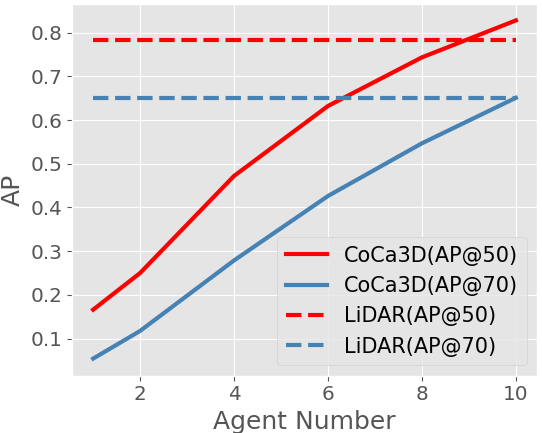}
    % \vspace{-2mm}
    \caption{OPV2V+}
    \label{fig:opv2v_agentnum}
  \end{subfigure}
  \begin{subfigure}{0.32\linewidth}
    \includegraphics[width=0.85\linewidth]{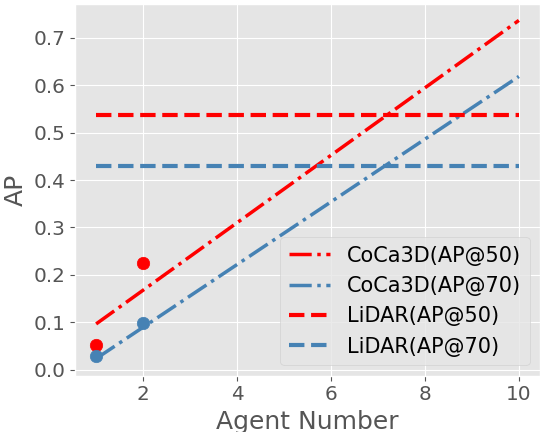}
    % \vspace{-2mm}
    \caption{DAIR-V2X}
    \label{fig:dair_agentnum}
  \end{subfigure}
  \begin{subfigure}{0.32\linewidth}
    \includegraphics[width=0.85\linewidth]{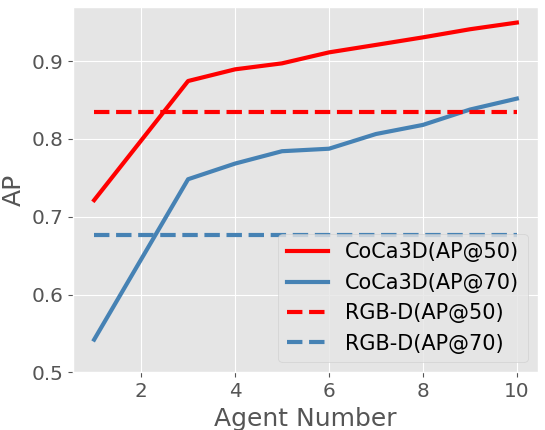}
    % \vspace{-2mm}
    \caption{CoPerception-UAVs+}
    \label{fig:uav_agentnum}
  \end{subfigure}
  \vspace{-3mm}
  \caption{CoCa3D steadily improves 3D detection performance as the number of agents grows. In OPV2V+, around $10$ collaborative agents enable collaborative cameras to catch up with LiDAR.}
  \vspace{-4mm}
  \label{Fig:abl_agentnumber}
\end{figure*}

\vspace{-2mm}
\section{Experimental Results}
\vspace{-1mm}

\begin{table*}[!t]
\centering
\caption{\texttt{CoCa3D} significantly outperforms the previous SOTAs, e.g. improve by 30.60\% on OPV2V, 12.59\% on CoPerception-UAVs, 44.21\% on DAIR-V2X for metric AP@70. }
% $*$ denotes the method with limited bandwidth. Comm is short for communication cost.
\label{tab:SOTAs}
\vspace{-3mm}
\begin{tabular}{l|ccc|ccc|ccc}
\hline
\multirow{2}{*}{Method} & \multicolumn{3}{c|}{OPV2V+} & \multicolumn{3}{c|}{CoPerception-UAVs+} & \multicolumn{3}{c}{DAIR-V2X}\\
       & AP@30 & AP@50  & AP@70 & AP@50  & AP@70  & AP@80 & AP@30 & AP@50  & AP@70\\ \hline
% LiDAR &   No Collaboration    & 0.8014 &  0.7839 & 0.6512 &- &- &- &- &- &- \\ \hline
% \multirow{8}{*}{Camera} 
No Collaboration     & 0.2748 & 0.2041 & 0.0853 & 0.6956 & 0.4900 & 0.2309 & 0.0977 & 0.0524 & 0.0305 \\
Late Fusion          & 0.6501 & 0.6198 & 0.5109 & 0.7206 & 0.5372 & 0.2597 & 0.2060 & 0.1078 & 0.0455 \\
When2com~\color{blue}{(\scriptsize{CVPR'20})}             & 0.4853 & 0.4211 & 0.3737 & 0.8219 & 0.6705 & 0.4102 &0.1957 &0.0984 &0.0459 \\
V2VNet~\color{blue}{(\scriptsize{ECCV'20})}                & 0.6246 & 0.5042 & 0.3852 & 0.9093 & 0.7177 & 0.3804 &0.1640 &0.0847 &0.0512 \\
DiscoNet~\color{blue}{(\scriptsize{NeurIPS'21})}              & 0.7300 & 0.6009 & 0.4179 & 0.9054 & 0.7079 & 0.3564 &0.1836 &0.1262 &0.0683 \\
V2X-ViT~\color{blue}{(\scriptsize{ECCV'22})}               & 0.8346 & 0.6659 & 0.3946 & 0.9094 & 0.7143 & 0.3525 &0.1862 &0.1075 &0.0490 \\
% $\text{Where2comm}^*$           & 0.8006 & 0.6783 & 0.4453 & -    & 0.8725 & 0.6091 & 0.2716 & 23.13\\
Where2comm~\color{blue}{(\scriptsize{NeurIPS'22})}            & 0.8191 & 0.7089 & 0.4741  & 0.9102 & 0.7383 & 0.3676 & 0.1754 & 0.1025 & 0.0547 \\
% Ours                 & - & - & - & -    & 0.8896 & 0.6149 & 0.2892 & 22.98 \\
% $\text{CoCa3D}^*$                & 0.8496 & 0.8081 & 0.6527  & 0.9342 & 0.7551 & 0.4375 & - & - & -  \\
CoCa3D                 & \textbf{0.8642} &  \textbf{0.8260} & \textbf{0.6675}   & \textbf{0.9497} & \textbf{0.8502} & \textbf{0.5835} & \textbf{0.3522} & \textbf{0.2260} & \textbf{0.0985} \\ 
\hline
\end{tabular}
\vspace{-6mm}
\end{table*}

\subsection{Datasets and experimental settings}
\vspace{-1mm}
We cover three datasets, both real-world and simulation scenarios, and multiple agent types. Metric Average Precision (AP) at Intersection-over-Union (IoU) threshold of $0.30$, $0.50$, $0.70$, and $0.80$ are used. Metric multi-class classification accuracy is used for depth accuracy.
The communication volume follows the standard setting as ~\cite{xu2022opv2v,xu2022v2xvit,hu2022where2comm} that counts the message size by byte in log scale with base $2$.
To compare communication results straightforwardly and fair, we do not consider any extra compression.

\noindent\textbf{OPV2V+.}
The original OPV2V~\cite{xu2022opv2v} is a large-scale vehicle-to-vehicle collaborative perception dataset, co-simulated by OpenCDA~\cite{xu2021opencda} and CARLA~\cite{dosovitskiy2017carla}. Here we introduce an extended version, OPV2V+, which  includes more collaborative agents ($\approx 10$). Each agent is equipt with 4 cameras and 4 depth sensors with resolution 600 $\times$ 800. The detection range is 281.6m $\times$ 80m. Our single-agent camera-only detector follows CaDDN~\cite{CaDDN} with 50 depth categories in linear-increasing spacing mode.

\noindent
\textbf{DAIR-V2X.} DAIR-V2X~\cite{yu2022dairv2x} is the only public \textbf{real-world} collaborative perception dataset. It contains two agents: a vehicle and a road-side-unit with image resolution 1080 $\times$ 1920. The perception range is 100m$\times$79m. Our single-agent camera-only detector also follows CaDDN~\cite{CaDDN}.

\noindent\textbf{CoPerception-UAVs+.}
The original CoPerception-UAVs~\cite{hu2022where2comm} is a large-scale UAV-based collaborative perception dataset, co-simulated by AirSim ~\cite{Airsim} and CARLA~\cite{dosovitskiy2017carla}. Here we introduce an extended version, CoPerception-UAVs+, which includes more collaborative agents ($\approx 10$). Each agent equips with 1 camera and 1 depth sensor with resolution 450 $\times$ 800. The detection range is 192m $\times$ 96m. Our single-agent camera-only detector follows DVDET~\cite{DVDET} with 10 depth categories in uniform spacing mode.

% We represent the field of view into a BEV map with size $(200, 504, 64)$ and the resolution is $0.4$m/pixel in length and width. 
\vspace{-2mm}
\subsection{Quantitative evaluation}
\vspace{-1mm}

\noindent\textbf{Collaborative camera-only 3D detection overtakes LiDAR.} 
Fig.~\ref{Fig:abl_agentnumber} investigates how the collaborative camera-only 3D detection performance changes with the number of collaborative agents on multiple datasets. To set up a goal, we consider a LiDAR-based detector implemented by the widely used PointPillar~\cite{lang2019pointpillars}.
Fig.~\ref{Fig:abl_agentnumber} (a) shows that: i) with $10$ collaborative agents, CoCa3D outperforms the LiDAR 3D detection at both AP@0.5/0.7 on OPV2V+! ii) the detection performance positively increases as the number of collaborative agents. And we then use the slope of OPV2V+ to fit the function of collaborative detection performance with agent number on DAIR-V2X as it only has two agents available. Fig.~\ref{Fig:abl_agentnumber} (b) shows that camera-only 3D detection is expected to outperform LiDAR 3D detection with $7$ collaborative vehicles in the real scenario.
As LiDAR is not a commonly available equipment for drones, we set a high-quality RGB-D camera instead.  
Fig.~\ref{Fig:abl_agentnumber} (c) shows that with $3$ collaborative drones, camera-only 3D detection outperforms ground-truth depth-based 3D detection. Furthermore, the steadily increasing detection performance with the number of collaborative agents encourages the agents to actively collaborate and achieve consistent improvement.

\noindent\textbf{Benchmark comparison.} Tab.~\ref{tab:SOTAs} compares the proposed \texttt{CoCa3D} with previous collaborative methods. We consider single-agent detection without collaboration (No collaboration), When2com~\cite{liu2020when2com}, V2VNet~\cite{wang2020v2vnet}, DiscoNet~\cite{li2021disconet}, V2X-ViT~\cite{xu2022v2xvit}, Where2comm~\cite{hu2022where2comm} and late fusion, where agents directly exchange the detected 3D boxes. We see that the proposed \texttt{CoCa3D} significantly outperforms previous state-of-the-arts, improves the SOTA performance by 30.60\% on OPV2V+, 12.59\% on CoPerception-UAVs+, 44.21\% on DAIR-V2X for AP@70. The reason is that collaborative depth estimation promotes more accurate 3D features per single agent, and the enhanced single agent features further facilitate collaborative detection feature learning, while previous collaborative methods do not specifically consider depth ambiguity and collaborate over single-agent features.

\begin{table}[!t]
\centering
\setlength\tabcolsep{2pt}
\caption{\texttt{CoCa3D} outperforms single-agent camera-only 3D detection with GT depth on OPV2V+ and CoPerception-UAVs+.}
\label{tab:OPV2V_upperbound}
\vspace{-2mm}
\begin{small}
\begin{tabular}{cc|ccc|ccc}
\hline
Co & Co & \multicolumn{3}{c|}{OPV2V+} & \multicolumn{3}{c}{CoPerception-UAVs+} \\
-Depth  & -FL& AP@30 & AP@50  & AP@70 & AP@50  & AP@70  & AP@80 \\\hline
% LiDAR        & - & - & 0.8014 &  0.7839 & 0.6512 & - & -& - \\ \hline
- & - & 0.2748 & 0.2041 & 0.0853 & 0.7213 & 0.5421 & 0.2846\\ 
 GT & - & 0.3454 & 0.2553 & 0.0973 & 0.8347 & 0.6764 & 0.4120\\
- & \checkmark & 0.8201 &  0.7191 & 0.4756 & 0.9084 & 0.7256 & 0.4028\\  % 0.82/0.71/0.47
GT & \checkmark & \textbf{0.9120} & \textbf{0.8805} & \textbf{0.7434} & \textbf{0.9505} & 0.8398 & 0.5504\\
\checkmark & \checkmark & 0.8642 & 0.8260 & 0.6675 & 0.9495 & \textbf{0.8518} & \textbf{0.5849} \\
\hline
\end{tabular}
\end{small}
\vspace{-3mm}
\end{table}

\noindent\textbf{Multi-agent collaboration evaluation.}
Tab.~\ref{tab:OPV2V_upperbound} assess the effectiveness of the proposed multi-agent collaboration on OPV2V+ and CoPerception-UAVs+ datasets. Ground-truth (GT) depth is included to provide the upper bound of collaborative camera-only 3D detection. We see that: i) both collaborations can consistently improve the performance; ii) camera-only 3D detection with collaboratively estimated depth significantly outperforms that with single-agent estimated depth and even approaches the upper bound with ground truth depth.

% \begin{table}[!ht]
% \centering
% \setlength\tabcolsep{2pt}
% \caption{\texttt{CoCa3D} outperforms single-agent camera-only 3D detection with ground-truth depth on CoPerception-UAVs.}
% \label{tab:UAV_upperbound}
% \vspace{-2mm}
% \begin{tabular}{cccccc}
% \hline
% Co-FL & Co-Depth & AP@50 & AP@70 & AP@80 & Acc \\ \hline
% % & - & N-Est & 0.6956 & 0.4900 & 0.2309 \\ 
% - & - & 0.7213 & 0.5421 & 0.2846 & 0.7586\\ 
% - & GT & 0.8347 & 0.6764 & 0.4120 & -\\
% % & \checkmark & N-Est & 0.8734 & 0.6686 & 0.3375 \\
% \checkmark & - & 0.9084 & 0.7256 & 0.4028 & 0.7586\\ 
% \checkmark & GT & \textbf{0.9505} & 0.8398 & 0.5504 &-\\
% \checkmark & \checkmark & 0.9495 & \textbf{0.8518} & \textbf{0.5849} & 0.8133\\\hline
% \end{tabular}
% \vspace{-3mm}
% \end{table}

\subsection{Qualitative evaluation}

\begin{figure*}[!t]
    \centering
    \vspace{2mm}
    % origin
    \begin{subfigure}[b]{0.19\textwidth}
        \centering
        \includegraphics[width=\textwidth]{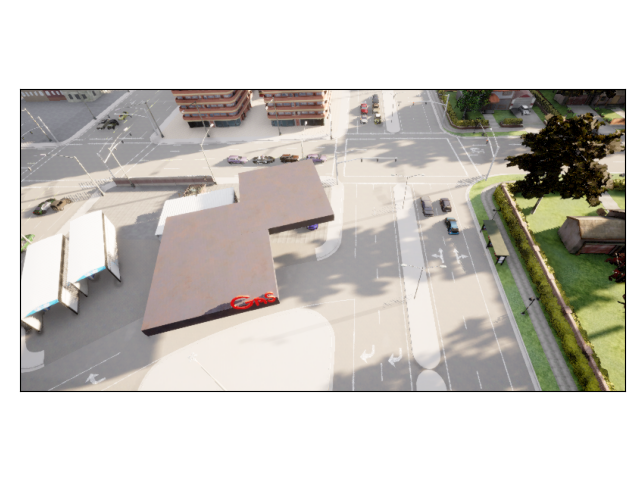}
        \caption{Image (RV)}
        \label{fig:uav_img}
    \end{subfigure}
    \hfill
    \begin{subfigure}[b]{0.19\textwidth}
        \centering
        \includegraphics[width=\textwidth]{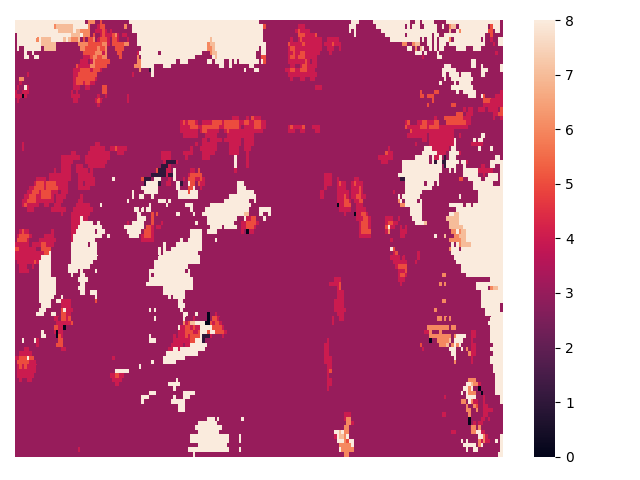}
        \caption{Single (RV)}
        \label{fig:uav_est}
    \end{subfigure}
    \hfill
    \begin{subfigure}[b]{0.19\textwidth}
        \centering
        \includegraphics[width=\textwidth]{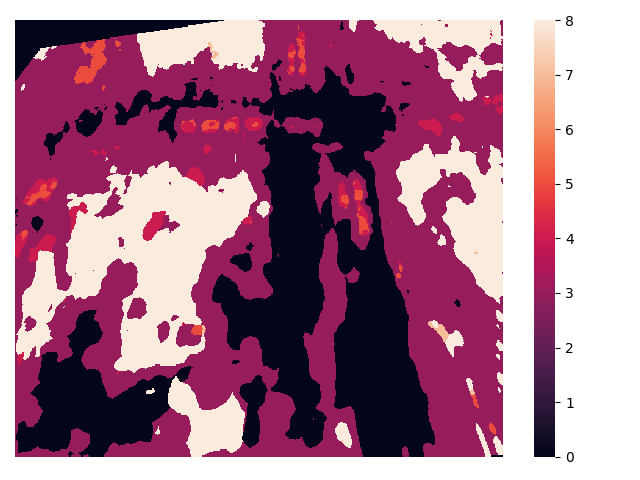}
        \caption{Collaboration (RV)}
        \label{fig:uav_colla_est}
    \end{subfigure}
    \hfill
    \begin{subfigure}[b]{0.19\textwidth}
        \centering
        \includegraphics[width=\textwidth]{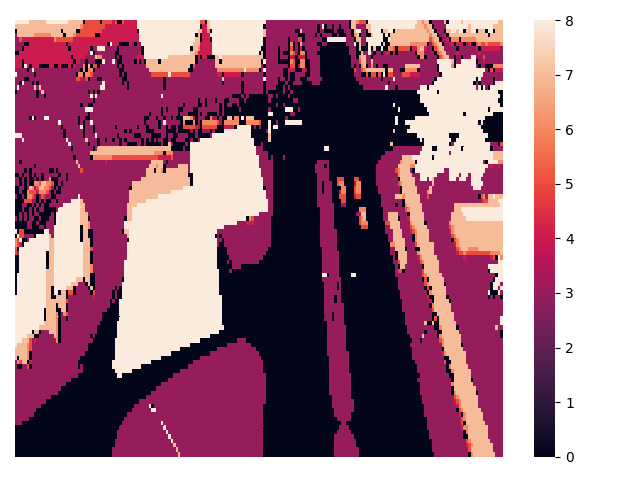}
        \caption{Ground-truth (RV)}
        \label{fig:uav_gt}
    \end{subfigure}
    \hfill
    \begin{subfigure}[b]{0.19\textwidth}
        \centering
        \includegraphics[width=\textwidth]{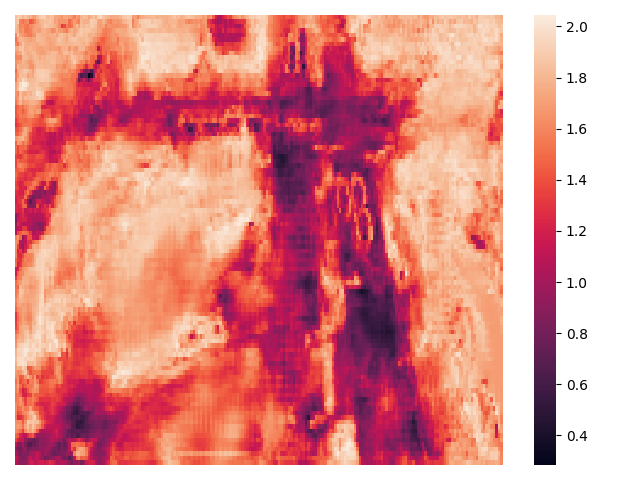}
        \caption{Uncertainty (RV)}
        \label{fig:uav_uncertainty}
    \end{subfigure}
    
    \centering
    \begin{subfigure}[b]{0.19\textwidth}
        \centering
        \includegraphics[width=\textwidth]{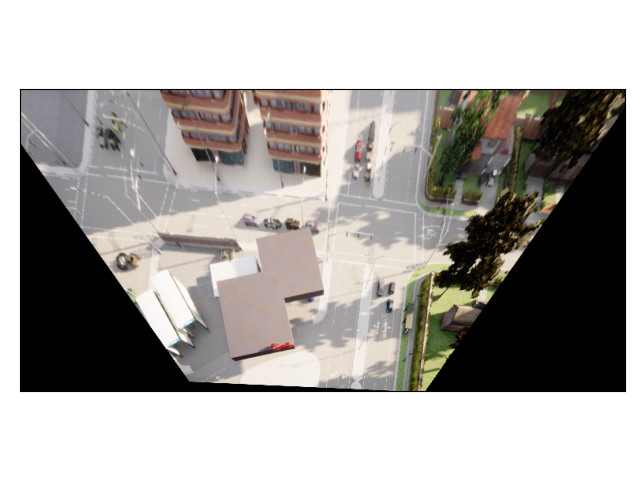}
        \caption{Image (BEV)}
        \label{fig:bev_img}
    \end{subfigure}
    \hfill
    \begin{subfigure}[b]{0.19\textwidth}
        \centering
        \includegraphics[width=\textwidth]{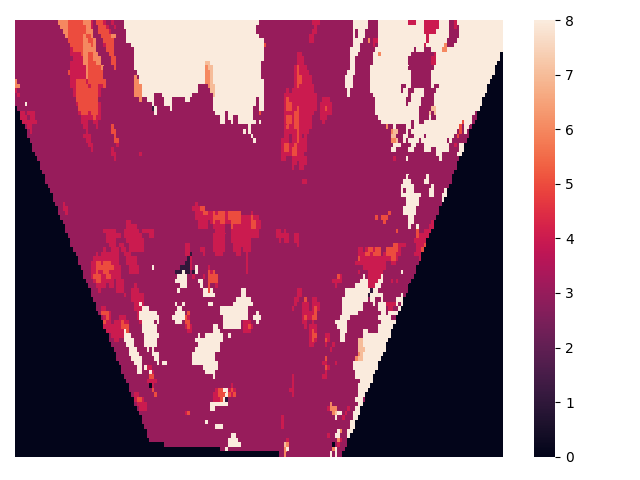}
        \caption{Single (BEV)}
        \label{fig:bev_est}
    \end{subfigure}
    \hfill
    \begin{subfigure}[b]{0.19\textwidth}
        \centering
        \includegraphics[width=\textwidth]{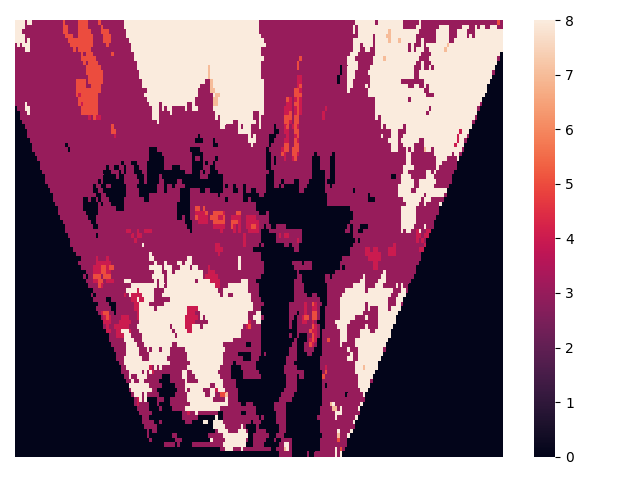}
        \caption{Collaboration (BEV)}
        \label{fig:bev_colla_est}
    \end{subfigure}
    \hfill
    \begin{subfigure}[b]{0.19\textwidth}
        \centering
        \includegraphics[width=\textwidth]{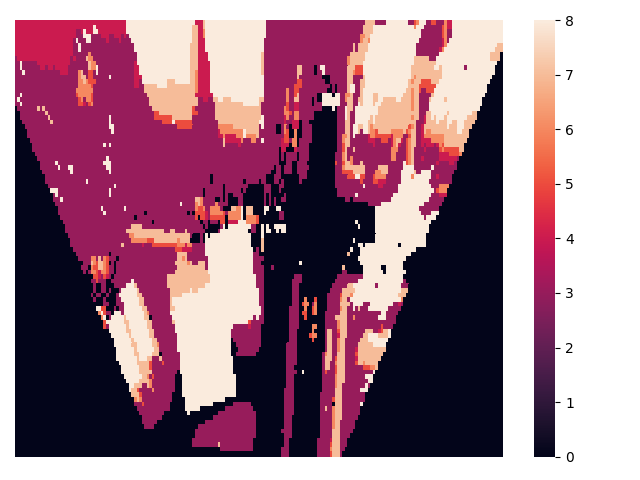}
        \caption{Ground-truth (BEV)}
        \label{fig:bev_gt}
    \end{subfigure}
    \hfill
    \begin{subfigure}[b]{0.19\textwidth}
        \centering
        \includegraphics[width=\textwidth]{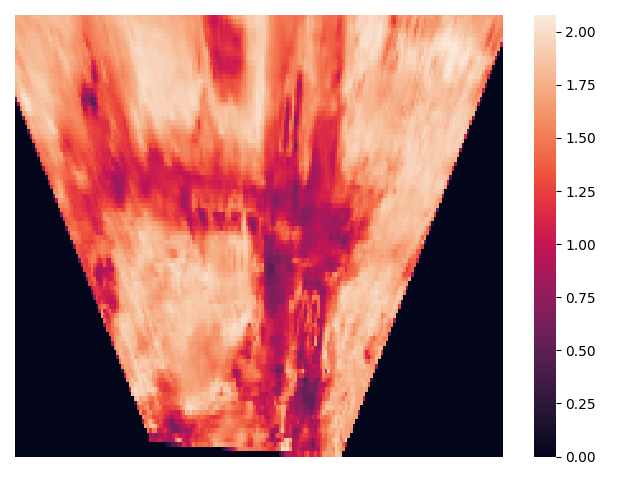}
        \caption{Uncertainty (BEV)}
        \label{fig:bev_uncertainty}
    \end{subfigure}
\vspace{-2mm}
\caption{Visualization of the depth and uncertainty in image range-view (RV) and bird's eye view (BEV) on CoPerception-UAVs+. Collaboratively estimated depth improves the single-agent estimated depth and approaches ground-truth depth.}
\label{fig:depth_u}
\vspace{-3mm}
\end{figure*}

\noindent\textbf{Visualization of depth and uncertainty.}
Fig.~\ref{fig:depth_u} shows that collaborative depth estimation outperforms single-agent depth estimation and approaches the ground truth depth. We see that: i) the single-agent depth estimation can estimate the relative depth while not able to precisely localize the depth candidate, for example, the vehicles are higher than the plane it sits while this plane is not grounded to the right category; ii) by introducing multi-view geometry, the collaboratively estimated depth can smoothly and accurately ground the plane; iii) the depth uncertainty is larger for the long-range and background areas. The reason is that the distant regions are hard to localize as they occupy too few image pixels, and the background regions are hard to localize due to the texture-less surfaces.

\begin{figure*}[!ht]
    \begin{subfigure}[b]{0.32\textwidth}
        \centering
        \includegraphics[width=\textwidth]{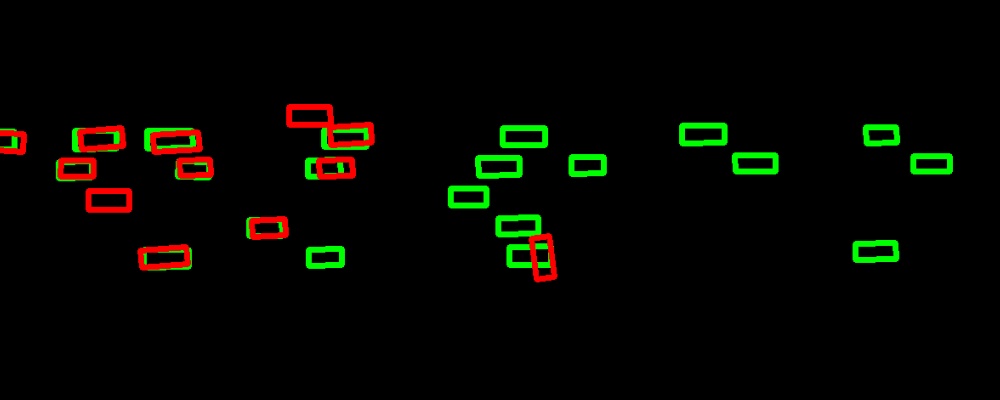}
        \caption{Camera}
        \label{fig:det_single_est}
    \end{subfigure}
    \hfill
    \begin{subfigure}[b]{0.32\textwidth}
        \centering
        \includegraphics[width=\textwidth]{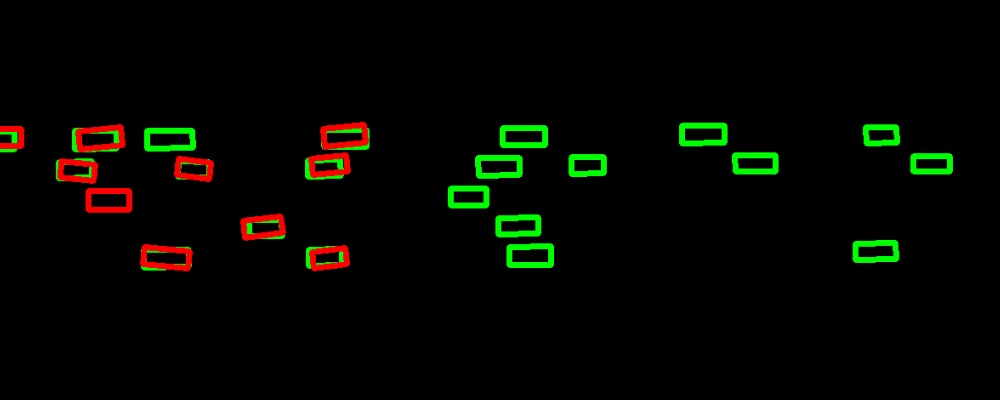}
        \caption{RGB-D}
        \label{fig:det_single_GT}
    \end{subfigure}
    \hfill
    \begin{subfigure}[b]{0.32\textwidth}
        \centering
        \includegraphics[width=\textwidth]{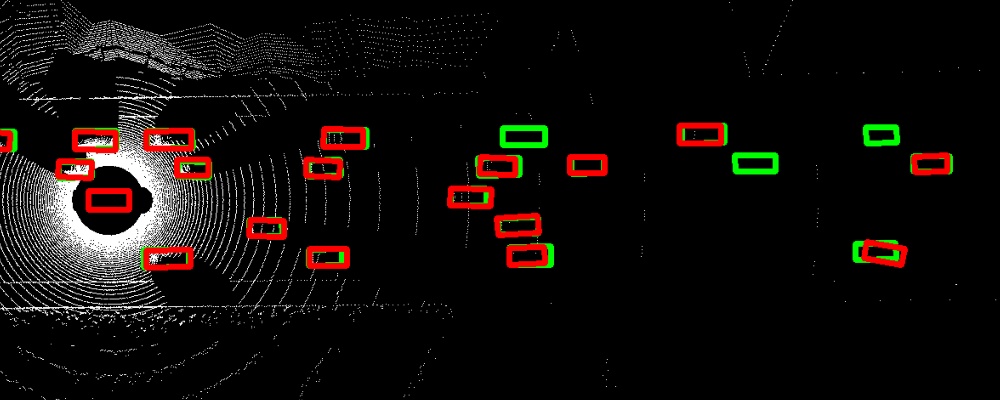}
        \caption{LiDAR}
        \label{fig:det_LiDAR}
    \end{subfigure}
    \hfill
    \begin{subfigure}[b]{0.32\textwidth}
        \centering
        \includegraphics[width=\textwidth]{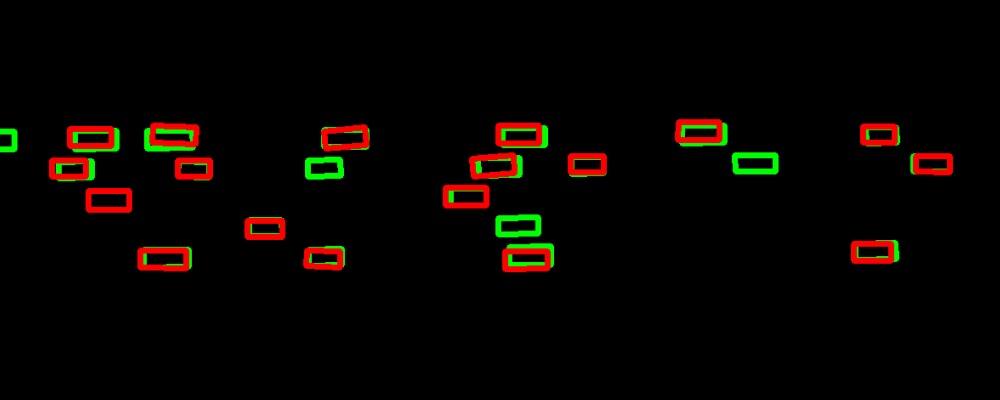}
        \caption{Camera with Co-FL}
        \label{fig:det_multi_est}
    \end{subfigure}
    \hfill
    \begin{subfigure}[b]{0.32\textwidth}
        \centering
        \includegraphics[width=\textwidth]{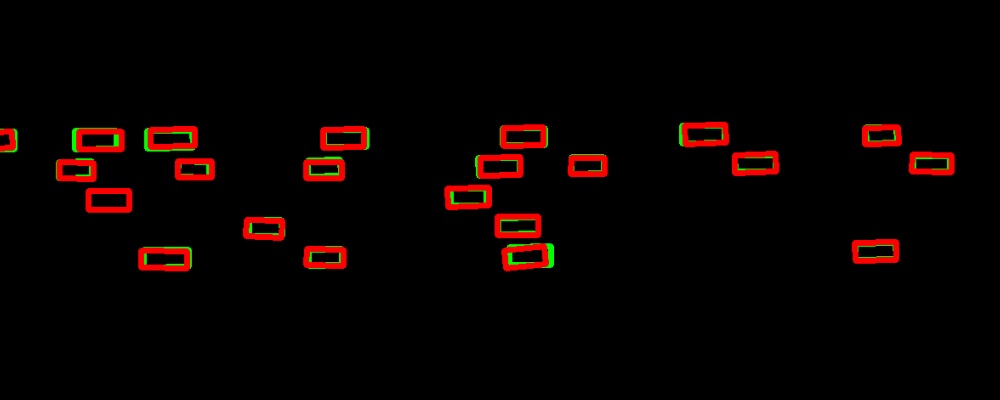}
        \caption{RGB-D with Co-FL}
        \label{fig:det_multi_gt}
    \end{subfigure}
    \hfill
    \begin{subfigure}[b]{0.32\textwidth}
        \centering
        \includegraphics[width=\textwidth]{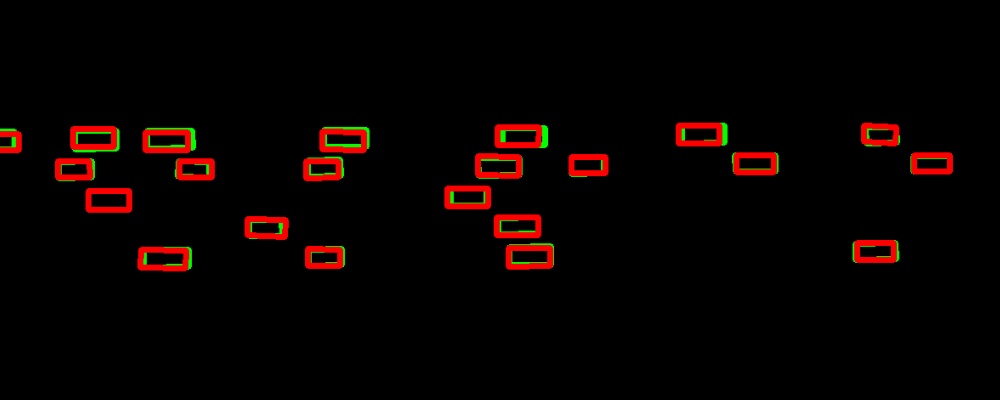}
        \caption{Camera with Co-Depth and Co-FL (\texttt{CoCa3D})}
        \label{fig:det_multi_coest}
    \end{subfigure}
\vspace{-2mm}
\caption{\texttt{CoCa3D} outperforms LiDAR detection on OPV2V+ with 10 agents. \textcolor{green}{Green} and \textcolor{red}{red} boxes denote GT and detection respectively.}
\label{fig:detections}
\vspace{-6mm}
\end{figure*}

\noindent\textbf{Visualization of detection results.}
Fig.~\ref{fig:detections} shows that compared to single LiDAR 3D detection, cameras with collaborations (Co-Depth and Co-FL) are able to achieve holistic and accurate detection results. The reason is that the single agent has some fundamental physical limitations such as long-range and occlusion issues. Fig.~\ref{fig:detections}(a-c) shows that augmenting a single agent sensor going from just a camera to including a depth sensor to LiDAR can improve the detection but still not able to achieve holistic detection. Fig.~\ref{fig:detections}(d)/(b) and (a)/(e) compares the detection results with and without collaborative feature learning, we see that collaboration can help detect lots of the missed objects in the single-agent detection. Fig.~\ref{fig:detections}(d,e,f) show that: i) both collaborations, Co-FL and Co-Depth, consistently improve the detection performance; ii) with collaboration, collaborative camera-only agents can outperform LiDAR in 3D detection.

\begin{figure}[!t]
    \centering
    \begin{subfigure}{0.49\linewidth}
    \includegraphics[width=0.99\linewidth]{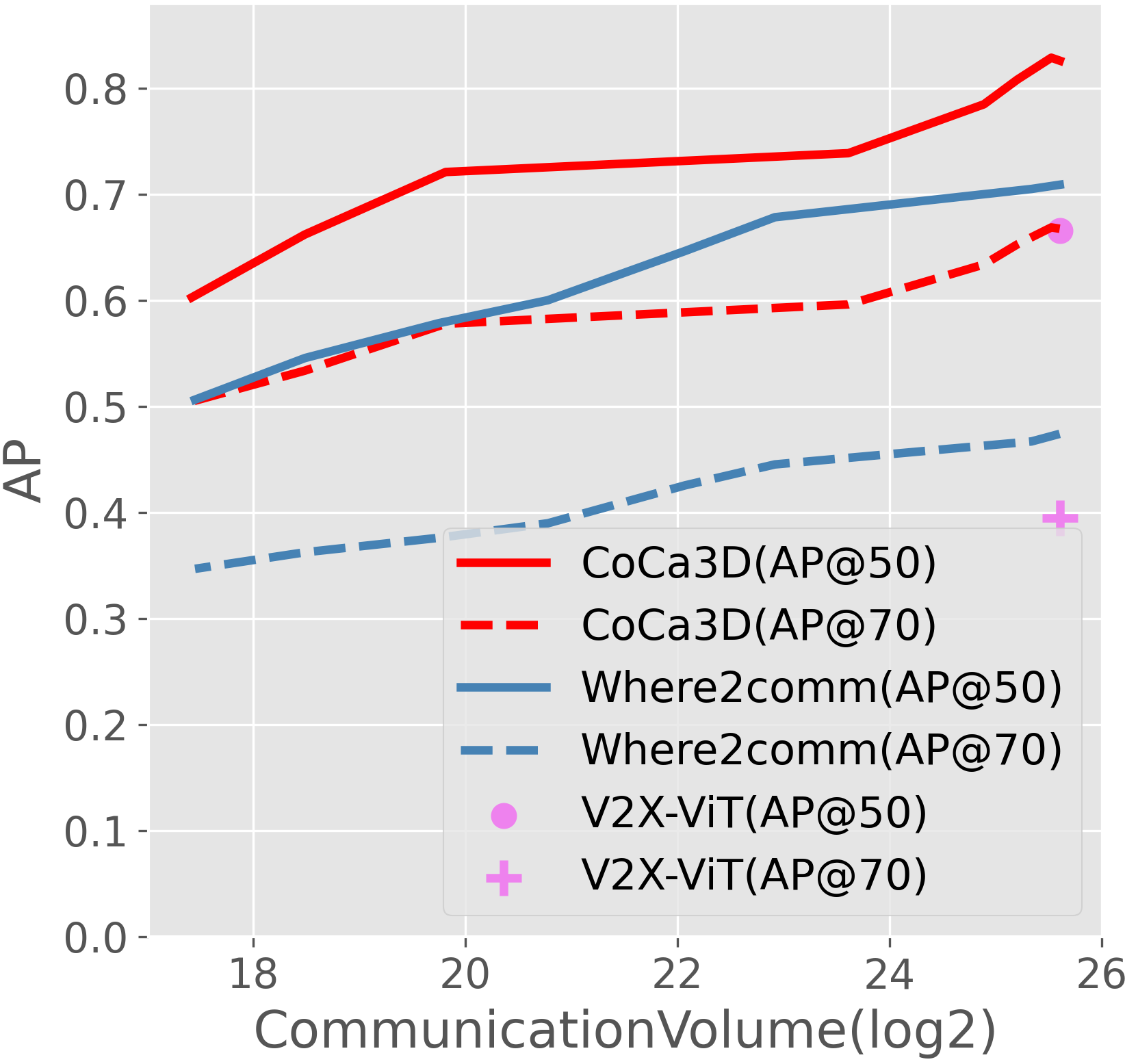}
    \vspace{-4mm}
    \caption{OPV2V+}
    \label{fig:opv2v_graph}
  \end{subfigure}
  \begin{subfigure}{0.49\linewidth}
    \includegraphics[width=0.99\linewidth]{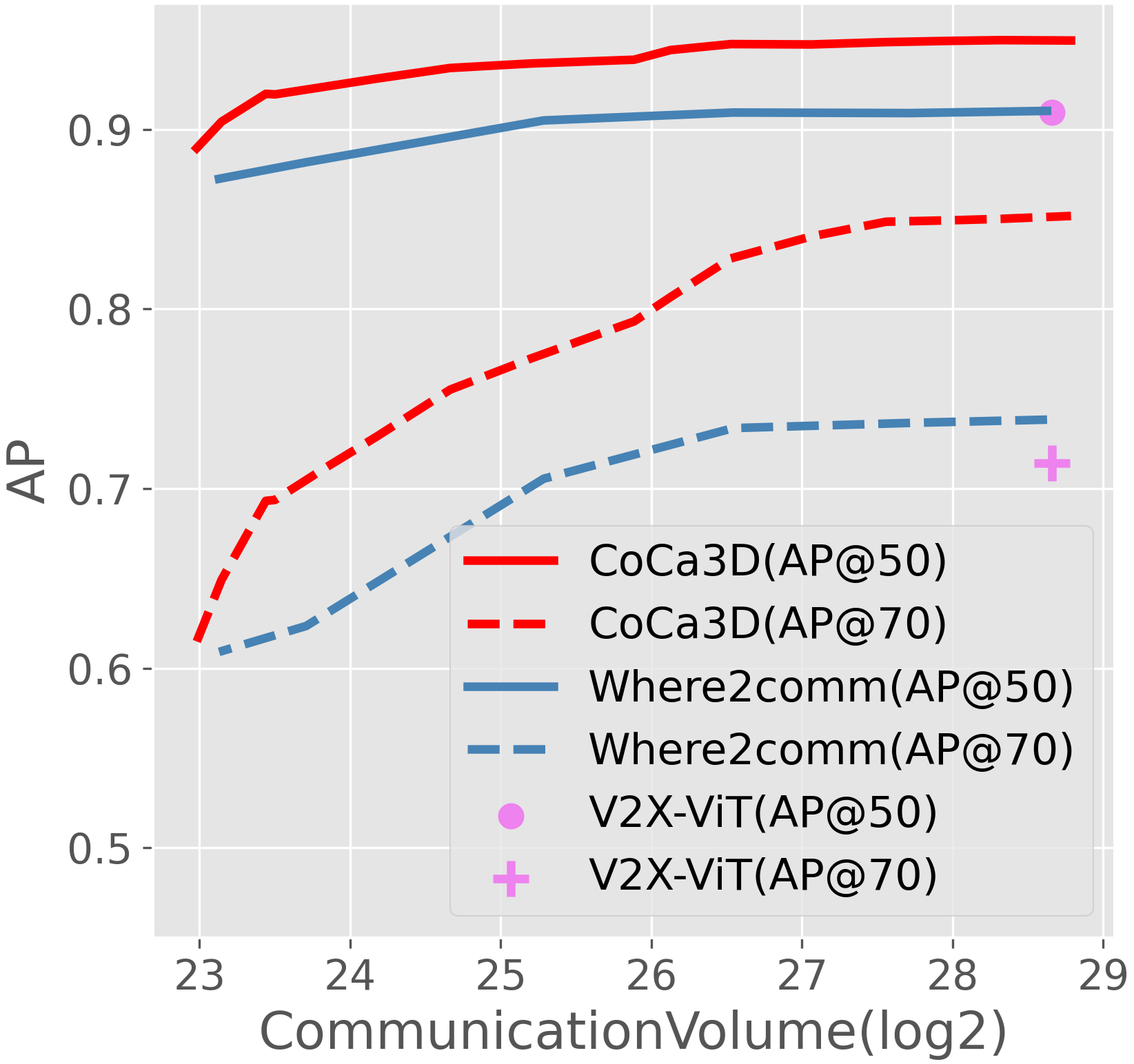}
    \vspace{-4mm}
    \caption{CoPerception-UAVs+}
    \label{fig:uav_graph}
  \end{subfigure}
  \vspace{-3mm}
  \caption{\texttt{CoCa3D} achieves superior detection performance and communication cost trade-off over various bandwidths.}
  \label{Fig:abl_tradeoff}
  \vspace{-3mm}
\end{figure}

\begin{table}[!t]
\setlength\tabcolsep{2pt}
\centering
\caption{Co-Depth significantly improves depth accuracy.}
\vspace{-2mm}
\label{tab:abl_depthacc}
\begin{small}
\begin{tabular}{ccc|ccc}
\hline
\multicolumn{3}{c|}{Full plane} & \multicolumn{3}{c}{Foreground} \\
Single & Collaboration & Gain & Single & Collaboration & Gain\\  \hline
0.6167 & 0.7781 & 26.17\% $\uparrow$ & 0.7586 &0.8133 & 7.21\% $\uparrow$ \\ \hline
\end{tabular}
\end{small}
\vspace{-2mm}
\end{table}

\begin{table}[!t]
\setlength\tabcolsep{2pt}
\centering
\caption{Ablation of collaborative depth estimation on CoPerception-UAVs+. Uniform/Linear denotes uniform/linear-increasing spacing. Dense and sparse supervision denote applying depth supervision over the full image plane and the object regions.}
\vspace{-2mm}
\label{tab:abl_codepth}
\begin{tabular}{cccccc}
\hline
Spacing & Supervision & AP@50  & AP@70 & AP@80 \\ \hline
Uniform & Dense & \textbf{0.9502} & 0.8444 & 0.5746 \\
Uniform & Sparse & 0.9330 & 0.8042 & 0.4829  \\
Linear & Dense  & 0.9484 &  0.8365 & 0.5580 \\
Linear & Sparse  & 0.9495 & \textbf{0.8518} & \textbf{0.5849} \\ \hline
\end{tabular}
\vspace{-3mm}
\end{table}

\vspace{-1mm}
\subsection{Ablation studies}
\vspace{-1mm}

\noindent\textbf{Trade-off between detection performance and communication cost.}
Fig.~\ref{Fig:abl_tradeoff} compares the proposed \texttt{CoCa3D} with the previous communication-efficient solution Where2comm~\cite{hu2022where2comm} in terms of the trade-off between detection performance (AP@IoU=50/70) and communication bandwidth. We see that: i) \texttt{CoCa3D} can adapt to varying bandwidths by adjusting the depth uncertainty threshold and the detection confidence threshold; ii) \texttt{CoCa3D} achieves superior detection performance and communication cost trade-off to Where2comm over varying communication bandwidth. The gain mainly comes from Co-Depth's improvement in depth estimation, which improves single-agent BEV features. With enhanced single BEV features, the Co-FL can generate better-augmented BEV features, resulting in improved detection performance.

\noindent\textbf{Effect of components in collaborative depth estimation.} Tab.~\ref{tab:abl_depthacc} assesses the gain of collaborative depth estimation (Co-Depth) over single-agent depth estimation (S-Depth) in the depth accuracy metric. We see that: i) Co-Depth steadily improves S-Depth over the full plane; ii) the depth accuracy of the foreground objects is higher than the full plane, as the texture-less backgrounds have fewer cues to localize. Tab.~\ref{tab:abl_codepth} assesses the effects of the depth spacing and supervision choices. We see that: i) overall, the Co-Depth is robust to the different spacing and supervision choices on AP@50/70; ii) linear-increasing spacing is robust to the various supervision choices that uniform spacing which fails the stricter metric AP@80 without dense supervision. The reason is that the linearly increasing spacing takes into account the prior depth distribution and assigns more depth candidates to frequently occurring depth ranges, which contributes to more balanced and easier depth learning.

%% file: contents/5-conclusion.tex
\section{Conclusion and Limitation}
% \vspace{-2mm}

We propose \texttt{CoCa3D}, a novel collaborative camera-only 3D detection that approaches holistic 3D detection. The core idea is to introduce multi-agent collaboration to improve the detection ability of cameras. Meanwhile, the communication cost is optimized, and each agent carefully selects the spatially sparse yet critical depth messages to share. Extensive experiments covering both real-world and simulation scenarios, and multi-type agents (cars, drones, and infrastructures) show that \texttt{CoCa3D} not only achieves state-of-the-art perception-bandwidth trade-off, but overtakes LiDAR-based detectors with a sufficient number of collaborative agents on OPV2V+.

\noindent
\textbf{Limitation and future work.} 
It is expensive to collect a real-world multi-agent perception dataset. So far, DAIR-V2X is the sole public real-world dataset, which only has one vehicle and one roadside unit. This work mainly leverages simulation data to validate the proposed novel methods and sketch a promising research direction. We advocate more resources for real-world data collection.

\noindent
\textbf{Acknowledgment.}
This research is supported by the National Key R\&D Program of China under Grant 2021ZD0112801, NSFC under Grant 62171276 and the Science and Technology Commission of Shanghai Municipal under Grant 21511100900 and 22DZ2229005.
\clearpage

%% file: contents/6-supp.tex
\section{Appendix}

% \subsection{LiDAR-based detector}

% \weidi{Here, we start by .... then .....}
Here, we start with dataset details, including generation and qualitative samples, then give out more implementation details.

\subsection{Datasets}
\subsubsection{OPV2V+}
\begin{figure}[!t]
    \centering
    \includegraphics[width=0.99\linewidth]{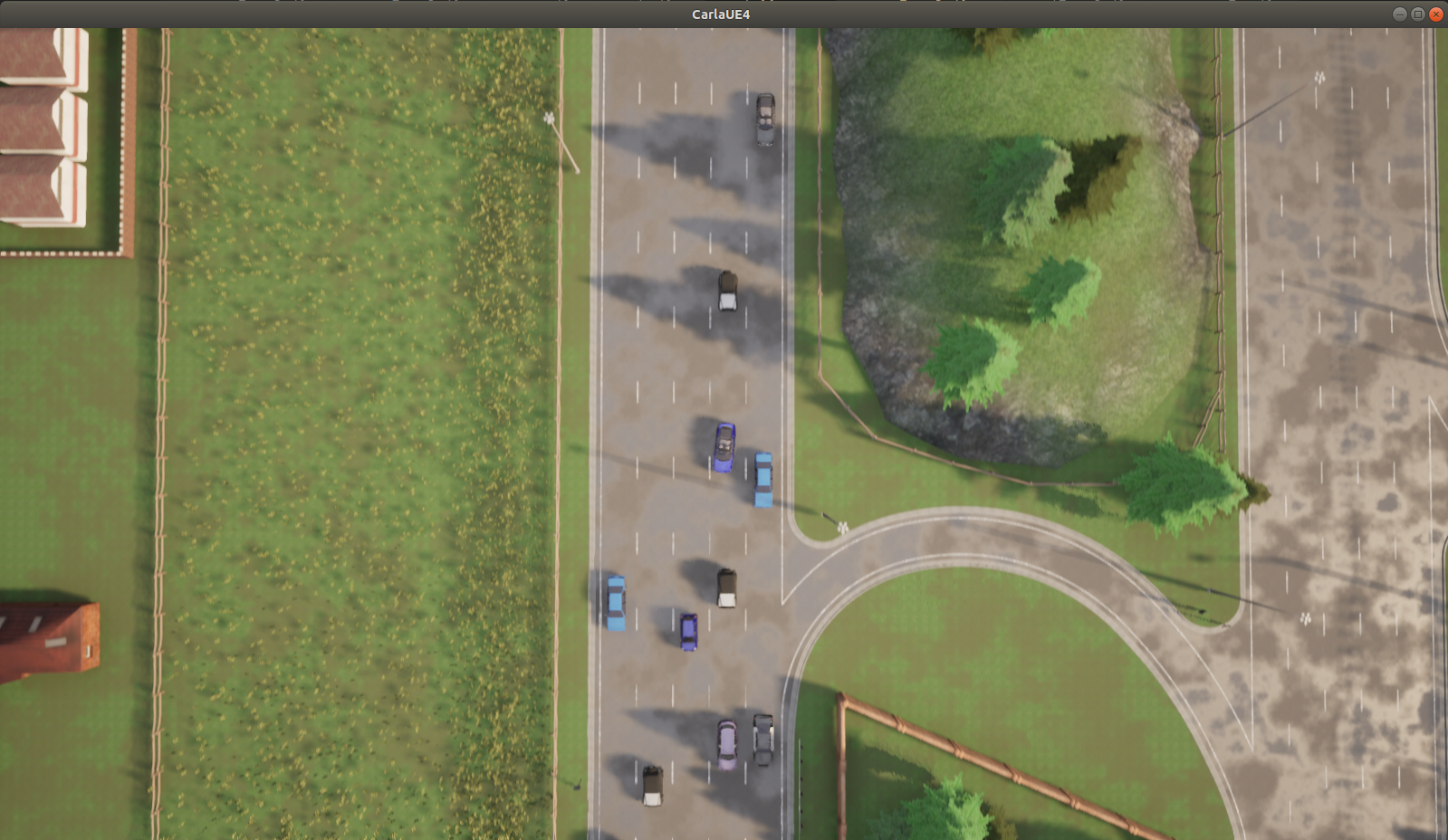}
  \vspace{-2mm}
  \caption{OPV2V+ is co-simulated by OpenCDA~\cite{xu2021opencda} and CARLA~\cite{dosovitskiy2017carla}.}
  \label{fig:opv2v_env}
  \vspace{-2mm}
\end{figure}

\begin{figure}[!t]
    \centering
    \begin{subfigure}{0.48\linewidth}
    \includegraphics[width=0.99\linewidth]{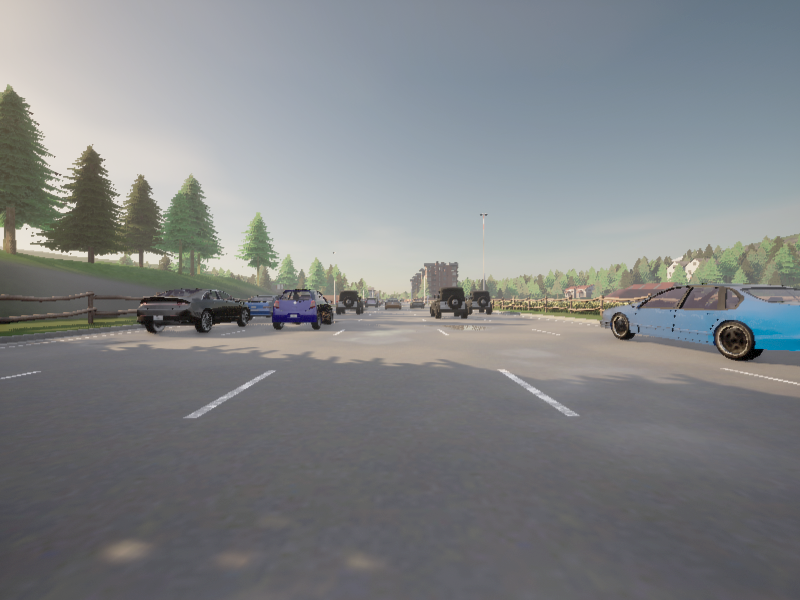}
    \caption{Camera 0}
  \end{subfigure}
  \hfill
  \begin{subfigure}{0.48\linewidth}
    \includegraphics[width=0.99\linewidth]{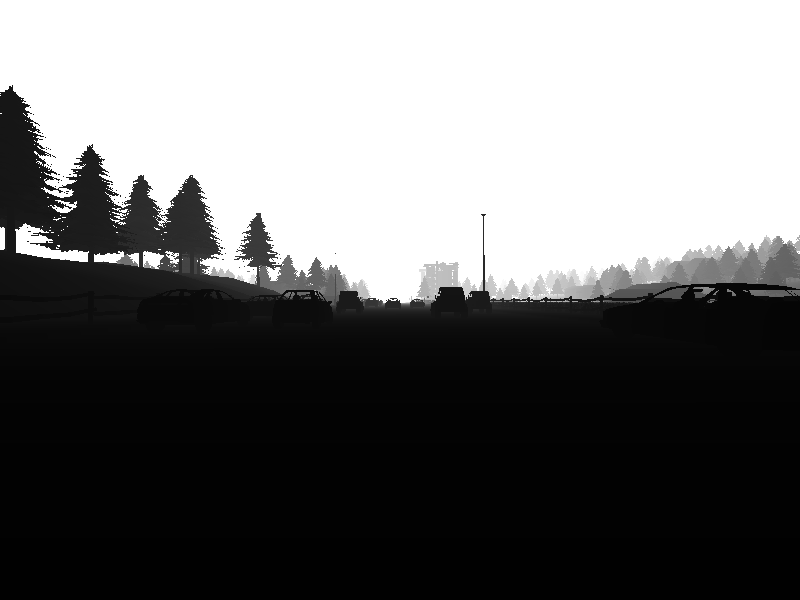}
    \caption{Depth 0}
  \end{subfigure}
  \hfill
  \begin{subfigure}{0.48\linewidth}
    \includegraphics[width=0.99\linewidth]{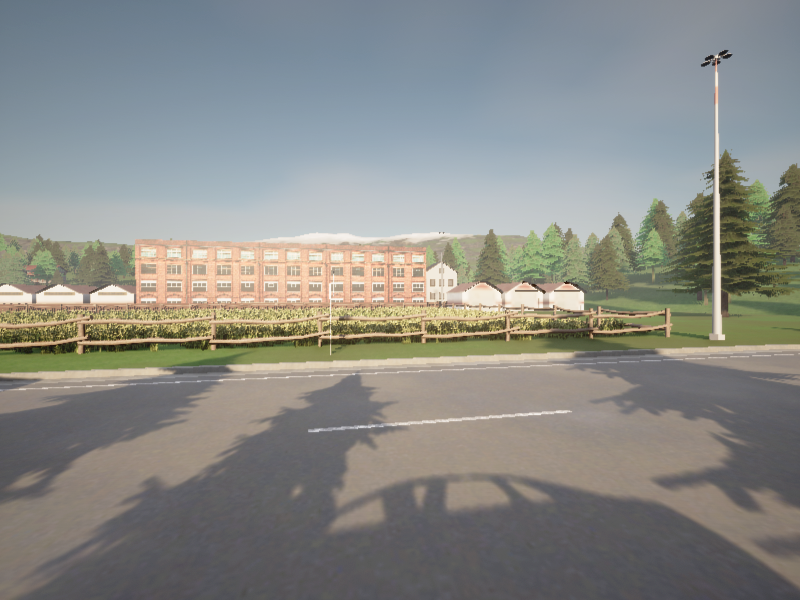}
    \caption{Camera 1}
  \end{subfigure}
  \hfill
  \begin{subfigure}{0.48\linewidth}
    \includegraphics[width=0.99\linewidth]{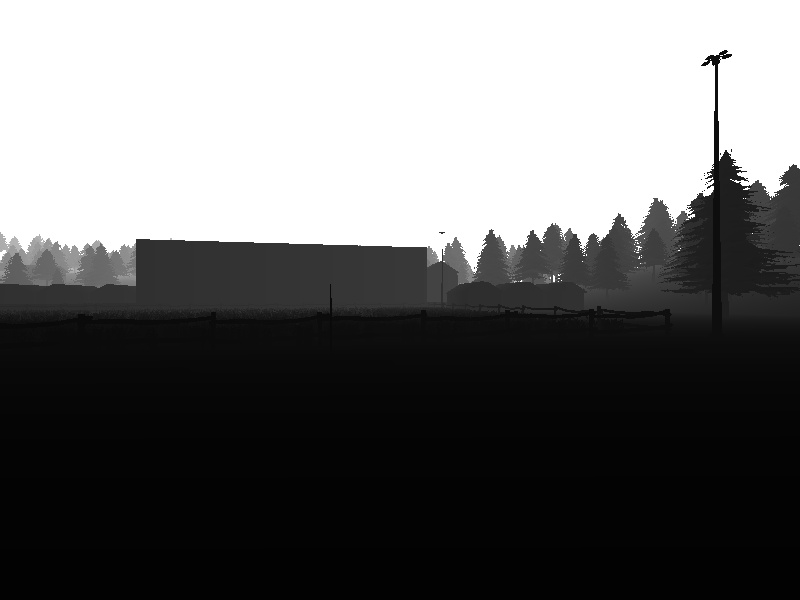}
    \caption{Depth 1}
  \end{subfigure}
  \hfill
  \begin{subfigure}{0.48\linewidth}
    \includegraphics[width=0.99\linewidth]{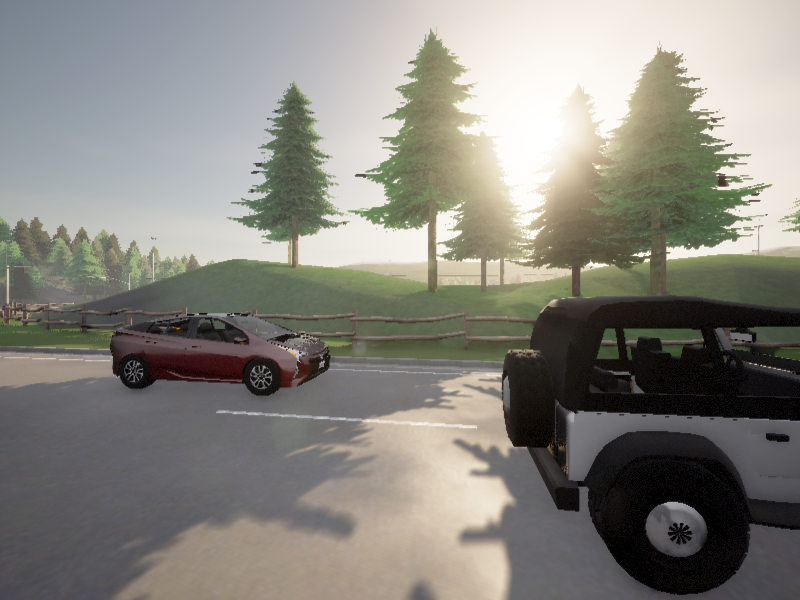}
    \caption{Camera 2}
  \end{subfigure}
  \hfill
  \begin{subfigure}{0.48\linewidth}
    \includegraphics[width=0.99\linewidth]{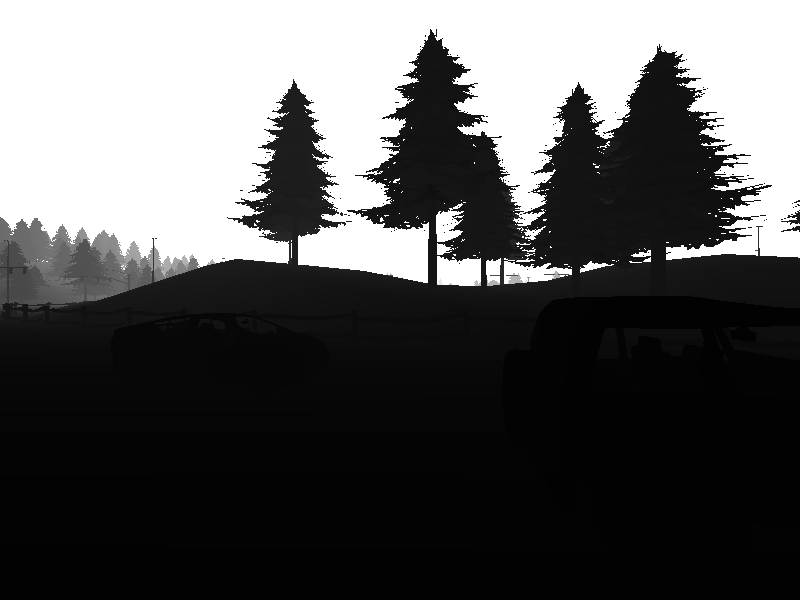}
    \caption{Depth 2}
  \end{subfigure}
  \hfill
  \begin{subfigure}{0.48\linewidth}
    \includegraphics[width=0.99\linewidth]{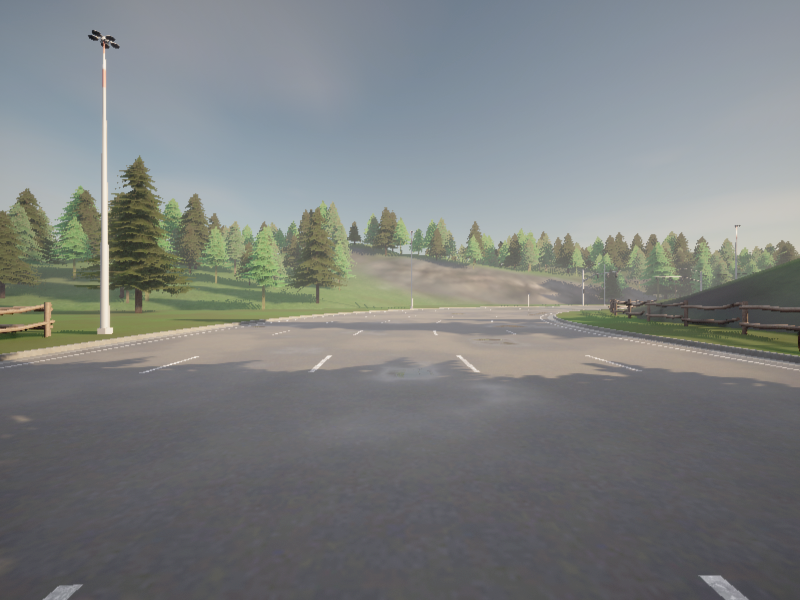}
    \caption{Camera 3}
  \end{subfigure}
  \hfill
  \begin{subfigure}{0.48\linewidth}
    \includegraphics[width=0.99\linewidth]{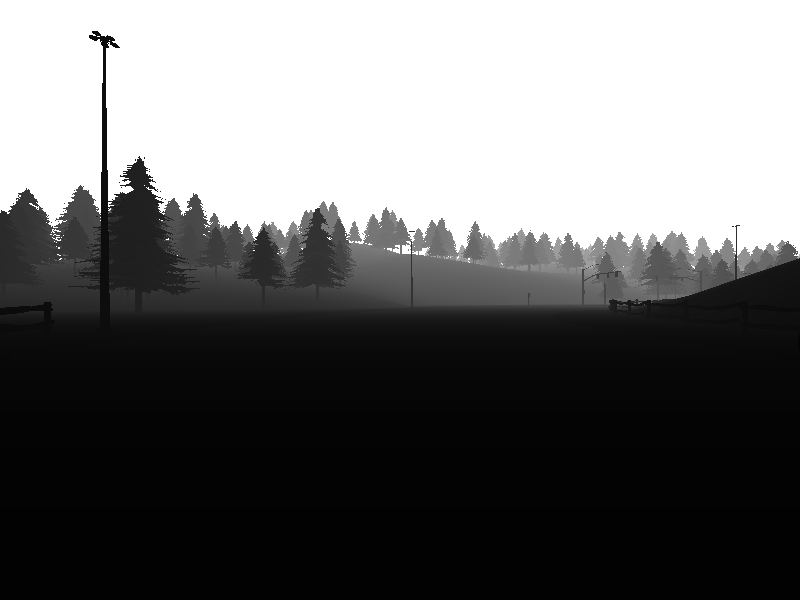}
    \caption{Depth 3}
  \end{subfigure}
  \vspace{-2mm}
  \caption{Each agent is equipped with 4 cameras and 4 depth sensors in OPV2V+.}
  \label{Fig:opv2v_single_sample}
  \vspace{-2mm}
\end{figure}

\begin{figure*}[!t]
  \centering
  \begin{subfigure}{0.24\linewidth}
    \includegraphics[width=0.99\linewidth]{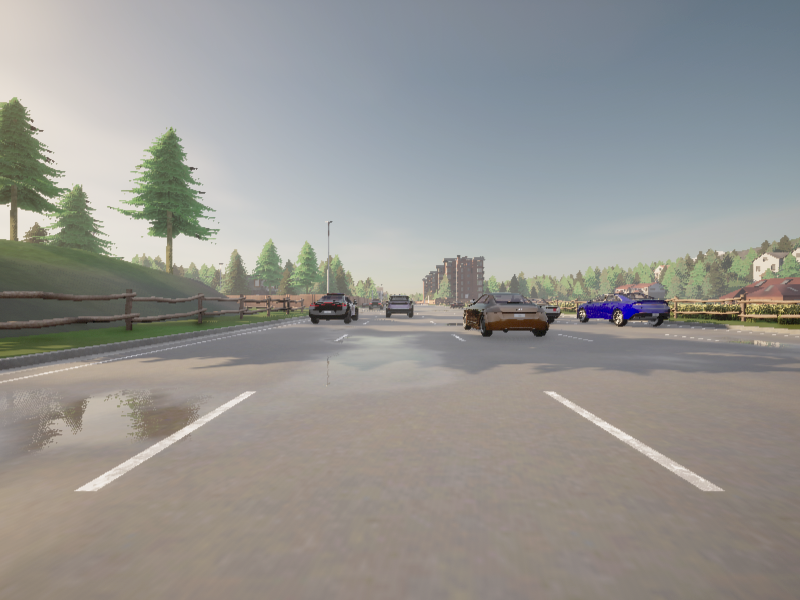}
    \caption{Agent 0: Camera 0}
  \end{subfigure}
  \hfill
  \begin{subfigure}{0.24\linewidth}
    \includegraphics[width=0.99\linewidth]{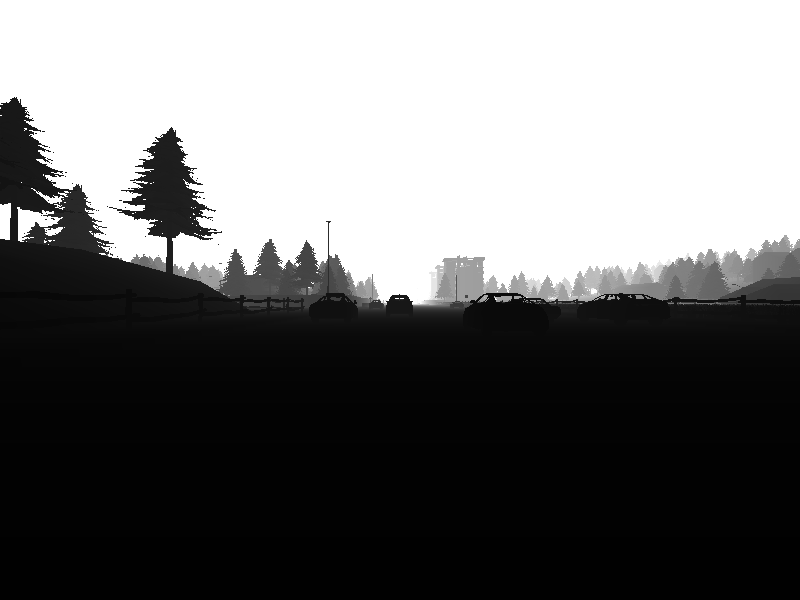}
    \caption{Agent 0: Depth 0}
  \end{subfigure}
  \hfill
  \begin{subfigure}{0.24\linewidth}
    \includegraphics[width=0.99\linewidth]{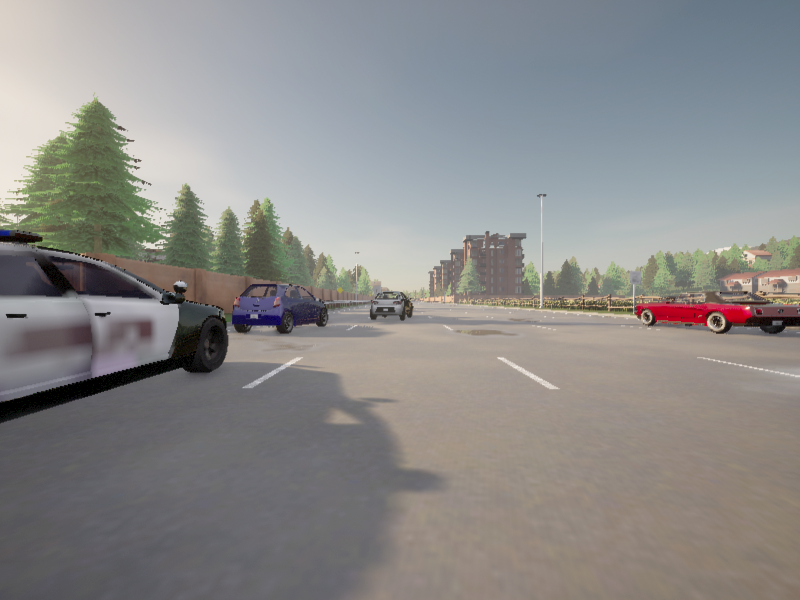}
    \caption{Agent 1: Camera 0}
  \end{subfigure}
  \hfill
  \begin{subfigure}{0.24\linewidth}
    \includegraphics[width=0.99\linewidth]{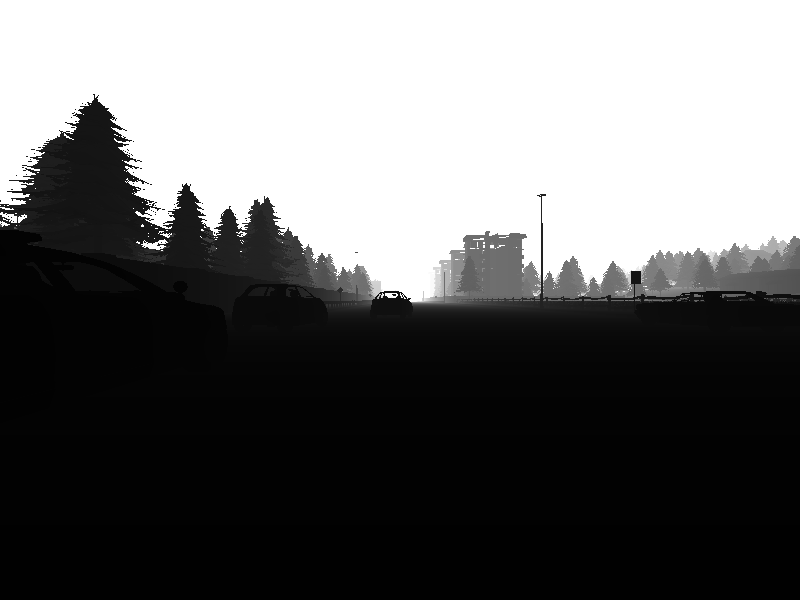}
    \caption{Agent 1: Depth 0}
  \end{subfigure}
  \hfill
  \begin{subfigure}{0.24\linewidth}
    \includegraphics[width=0.99\linewidth]{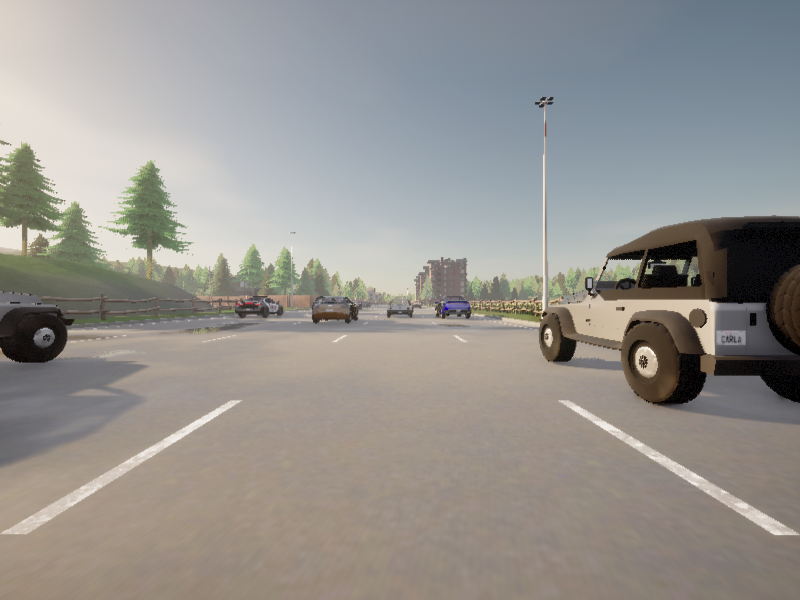}
    \caption{Agent 2: Camera 0}
  \end{subfigure}
  \hfill
  \begin{subfigure}{0.24\linewidth}
    \includegraphics[width=0.99\linewidth]{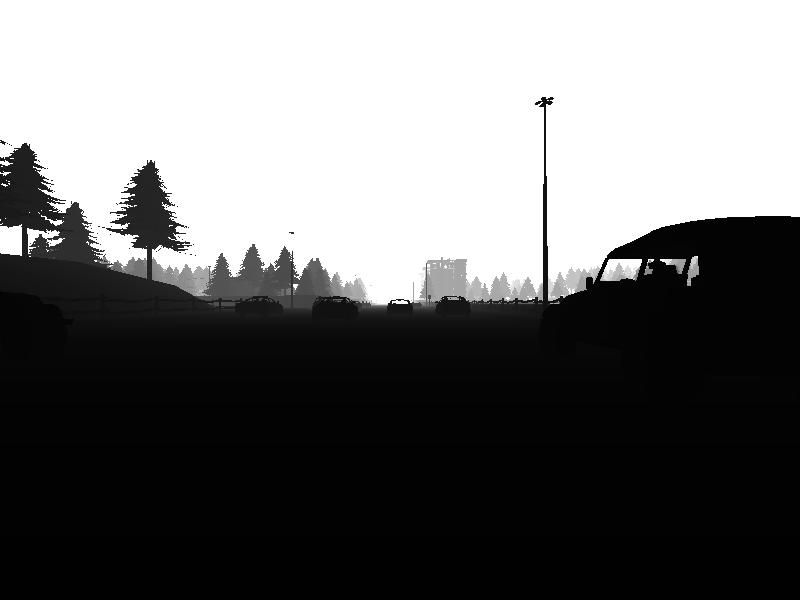}
    \caption{Agent 2: Depth 0}
  \end{subfigure}
  \hfill
  \begin{subfigure}{0.24\linewidth}
    \includegraphics[width=0.99\linewidth]{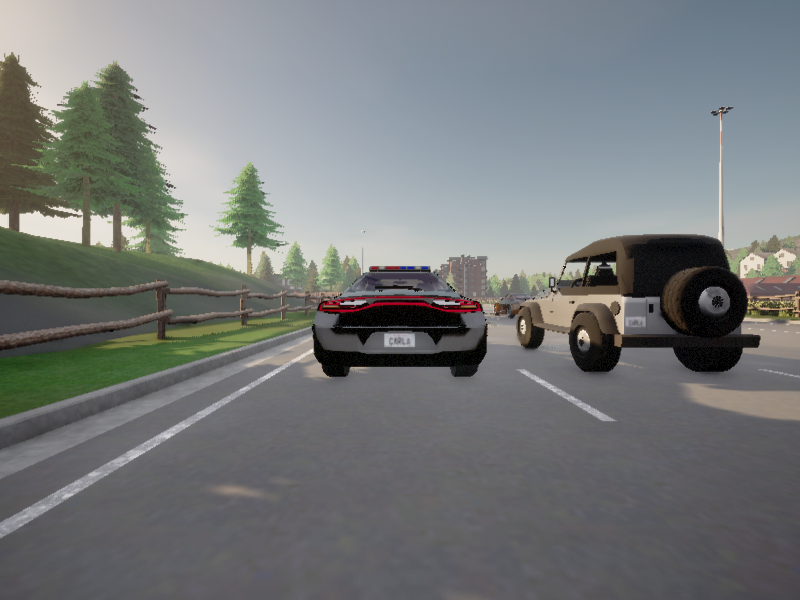}
    \caption{Agent 3: Camera 0}
  \end{subfigure}
  \hfill
  \begin{subfigure}{0.24\linewidth}
    \includegraphics[width=0.99\linewidth]{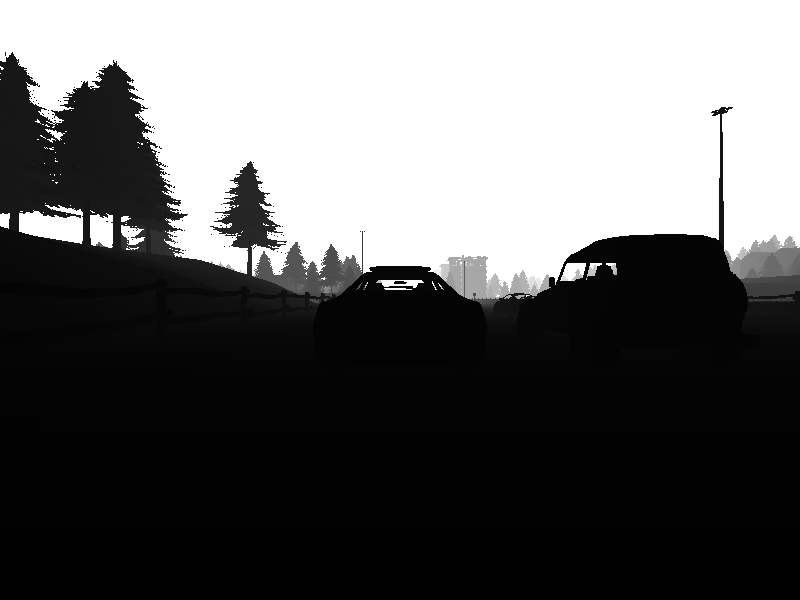}
    \caption{Agent 3: Depth 0}
  \end{subfigure}
  \hfill
  \begin{subfigure}{0.24\linewidth}
    \includegraphics[width=0.99\linewidth]{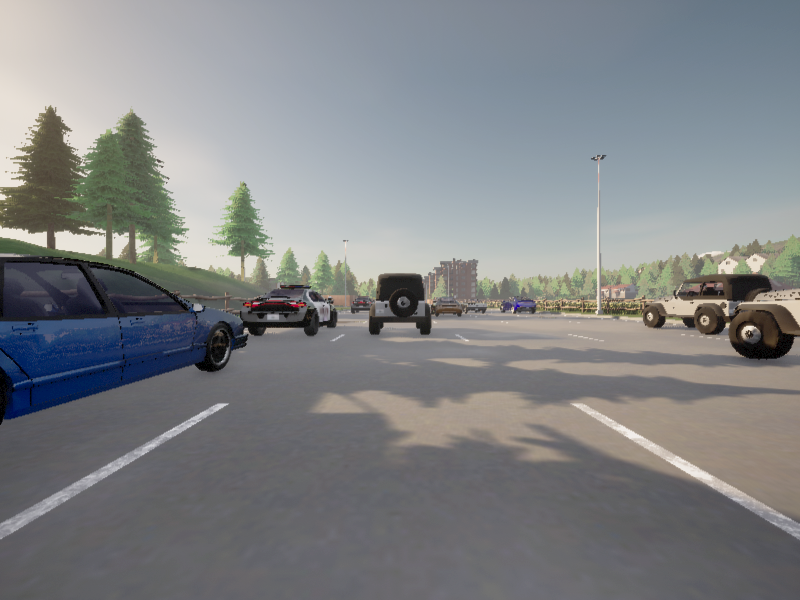}
    \caption{Agent 4: Camera 0}
  \end{subfigure}
  \hfill
  \begin{subfigure}{0.24\linewidth}
    \includegraphics[width=0.99\linewidth]{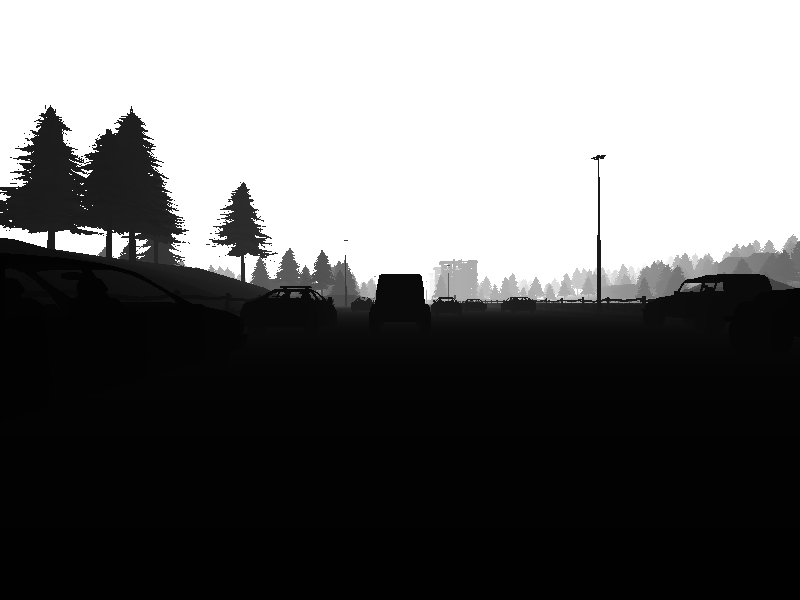}
    \caption{Agent 4: Depth 0}
  \end{subfigure}
  \hfill
  \begin{subfigure}{0.24\linewidth}
    \includegraphics[width=0.99\linewidth]{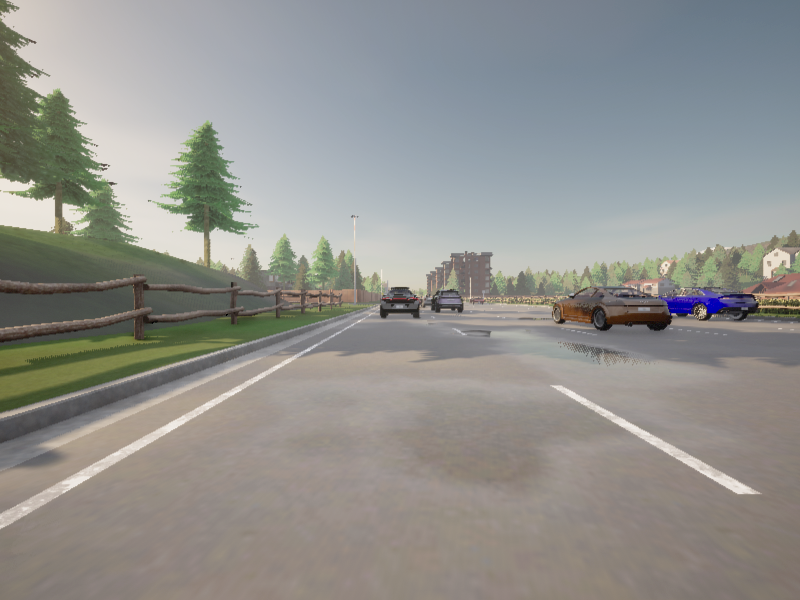}
    \caption{Agent 5: Camera 0}
  \end{subfigure}
  \hfill
  \begin{subfigure}{0.24\linewidth}
    \includegraphics[width=0.99\linewidth]{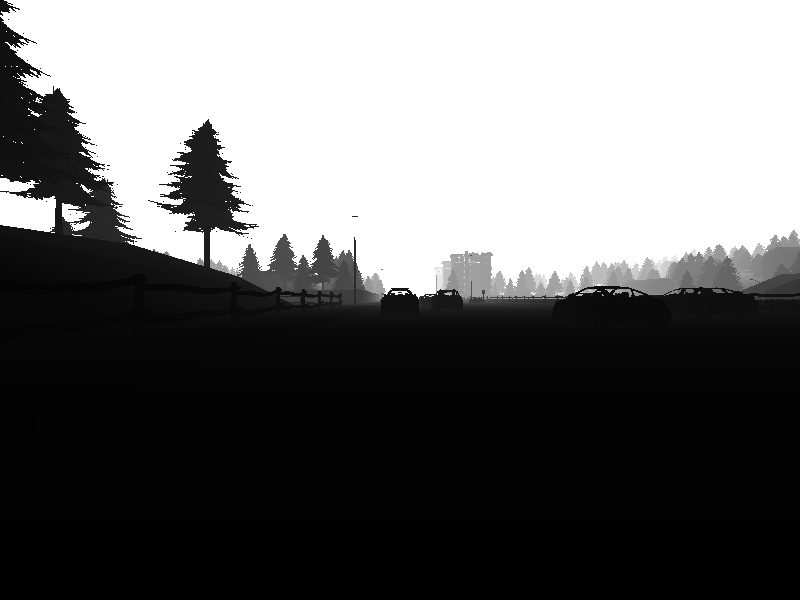}
    \caption{Agent 5: Depth 0}
  \end{subfigure}
  \hfill
  \begin{subfigure}{0.24\linewidth}
    \includegraphics[width=0.99\linewidth]{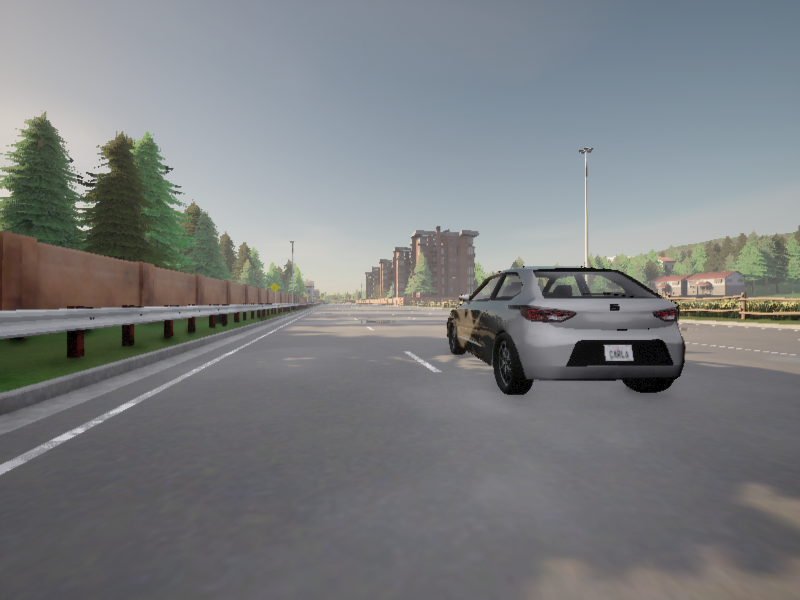}
    \caption{Agent 6: Camera 0}
  \end{subfigure}
  \hfill
  \begin{subfigure}{0.24\linewidth}
    \includegraphics[width=0.99\linewidth]{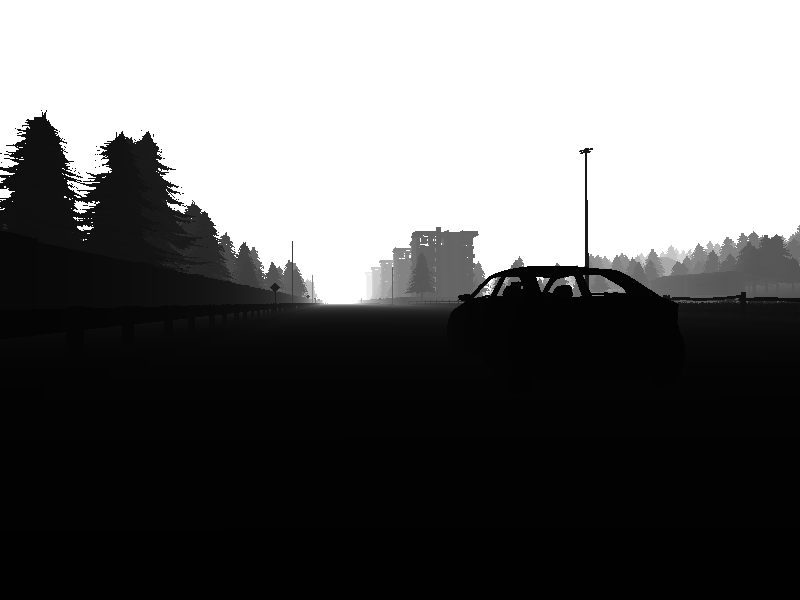}
    \caption{Agent 6: Depth 0}
  \end{subfigure}
  \hfill
  \begin{subfigure}{0.24\linewidth}
    \includegraphics[width=0.99\linewidth]{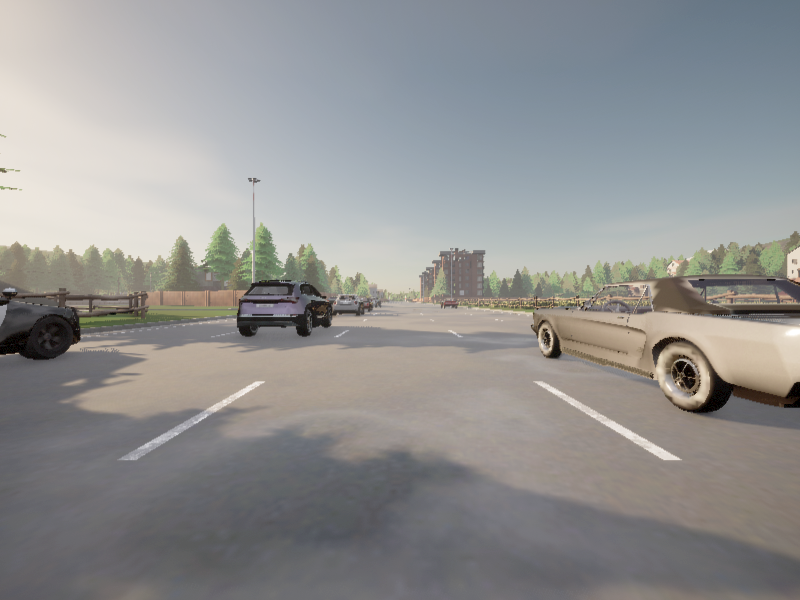}
    \caption{Agent 7: Camera 0}
  \end{subfigure}
  \hfill
  \begin{subfigure}{0.24\linewidth}
    \includegraphics[width=0.99\linewidth]{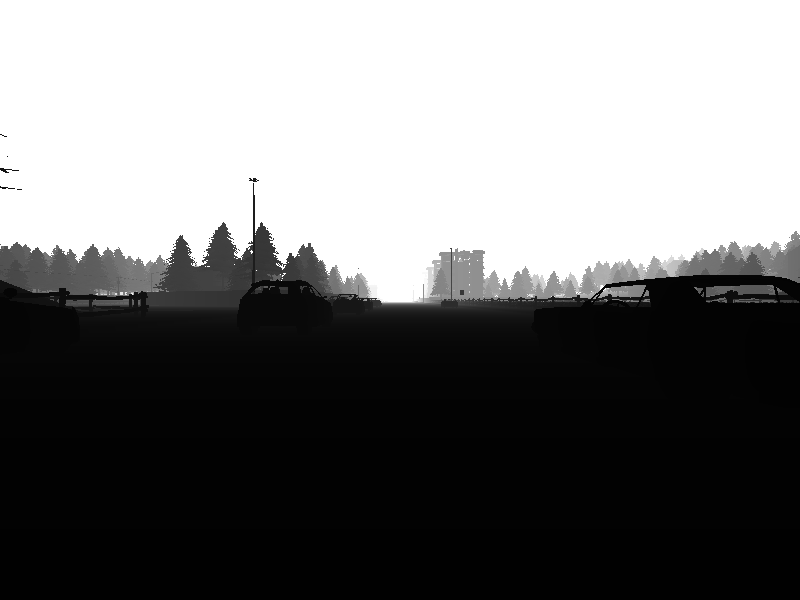}
    \caption{Agent 7: Depth 0}
  \end{subfigure}
  \hfill
  \begin{subfigure}{0.24\linewidth}
    \includegraphics[width=0.99\linewidth]{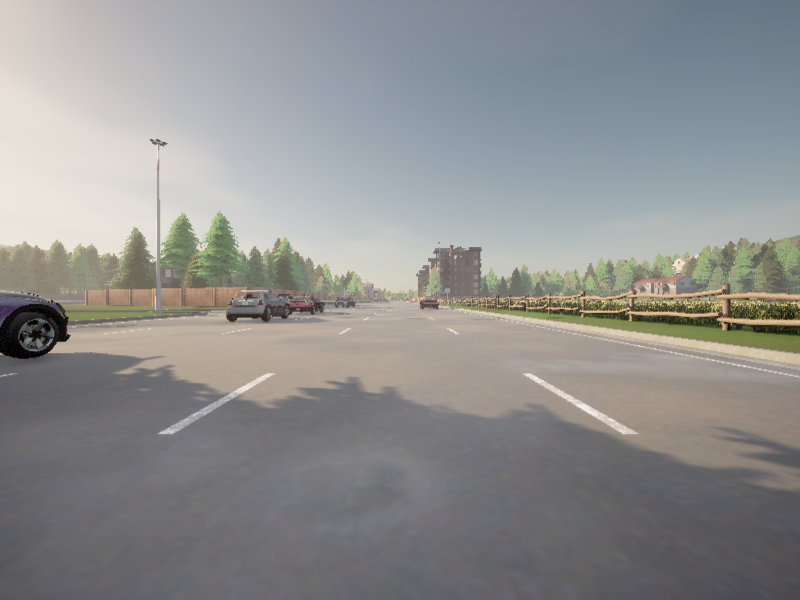}
    \caption{Agent 8: Camera 0}
  \end{subfigure}
  \hfill
  \begin{subfigure}{0.24\linewidth}
    \includegraphics[width=0.99\linewidth]{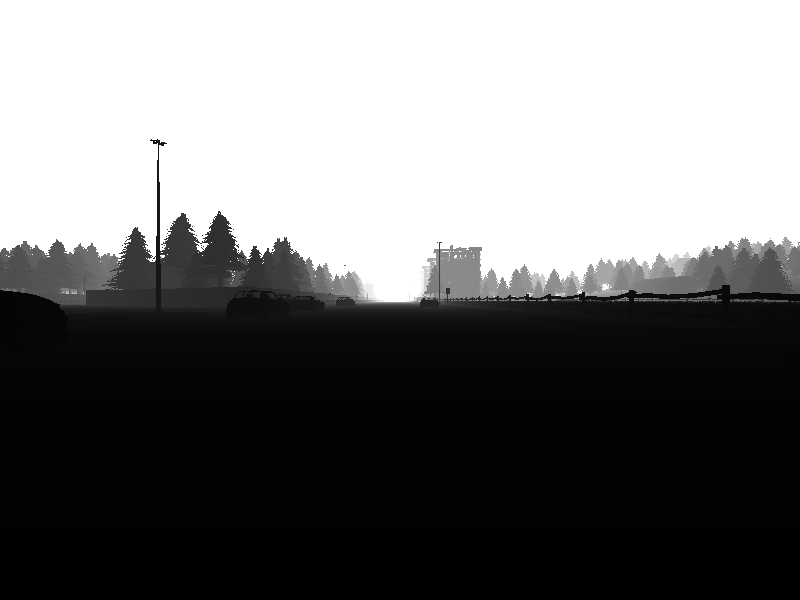}
    \caption{Agent 8: Depth 0}
  \end{subfigure}
  \hfill
  \begin{subfigure}{0.24\linewidth}
    \includegraphics[width=0.99\linewidth]{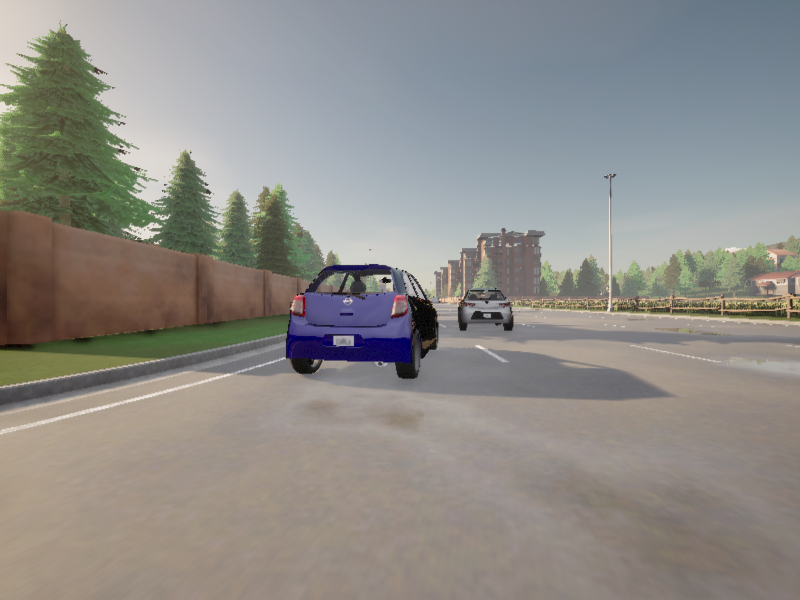}
    \caption{Agent 9: Camera 0}
  \end{subfigure}
  \hfill
  \begin{subfigure}{0.24\linewidth}
    \includegraphics[width=0.99\linewidth]{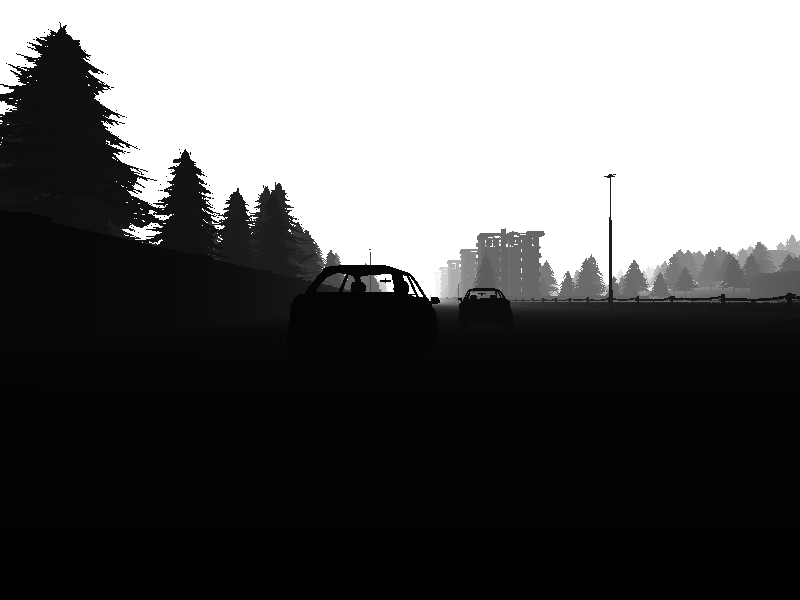}
    \caption{Agent 9: Depth 0}
  \end{subfigure}
  \vspace{-2mm}
  \caption{Data sample with 10 agents of OPV2V+.}
  \label{Fig:opv2v_sample}
  \vspace{-2mm}
\end{figure*}

\noindent
\textbf{Data generation.} We extend the original OPV2V~\cite{xu2022opv2v} with more collaborative agents ($10$). Our OPV2V+ is co-simulated by OpenCDA~\cite{xu2021opencda} and CARLA~\cite{dosovitskiy2017carla}. Figure.~\ref{fig:opv2v_env} shows the simulation environment. OpenCDA provides the driving scenarios which ensure the agents drive smoothly and safely, including the vehicle's initial location and moving speed. CARLA provides the maps, and weather and controls the movements of the agents. We replay the simulation logs of OPV2V and equip more vehicles with camera and depth sensors. Figure.~\ref{Fig:opv2v_single_sample} shows the four views (front, left, right, back) of the same agent. Figure.~\ref{Fig:opv2v_sample} shows a randomly selected data sample with 10 collaborative agents, the collected front view images in the same timestamp.

\begin{figure}[!h]
\centering
\noindent
% \hspace{1mm}
\centering
    \begin{subfigure}{0.7\linewidth}
    \includegraphics[width=0.99\linewidth]{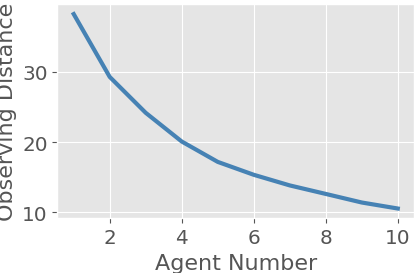}
    % \vspace{-6mm}
    \caption{Distance (1-10)}
    \label{fig:dis_distribution}
  \end{subfigure}
  \begin{subfigure}{0.8\linewidth}
    \includegraphics[width=0.99\linewidth]{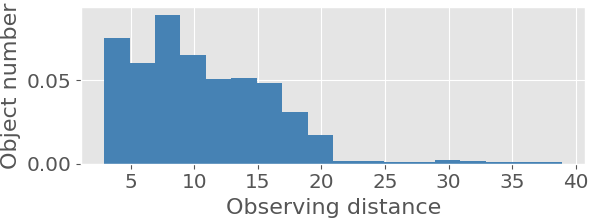}
    % \vspace{-6mm}
    \caption{Distance histogram(10 agents)}
    \label{fig:dis_hist}
  \end{subfigure}
    \begin{subfigure}{0.8\linewidth}
    \includegraphics[width=0.99\linewidth]{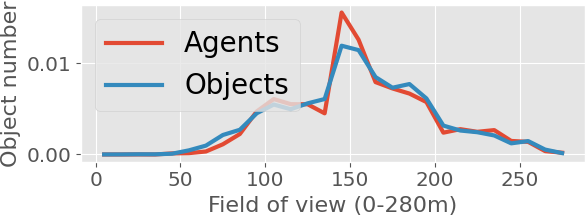}
    % \vspace{-6mm}
    \caption{Agent distribution(10 agents)}
    \label{fig:agent_hist}
  \end{subfigure}
  % \vspace{-3mm}
  \caption{Object distance distribution.}
\label{fig:agent_dist}
\vspace{-6mm}
\end{figure}

\noindent
\textbf{Agent distribution. } We provide more statistical analysis of the distance distribution between agents to objects. 1) The distance between objects and their closest agents decreases as the number of agents increases, see Fig.~\ref{fig:dis_distribution}. Given 10 agents, this distance is mostly within 20m, see Fig.~\ref{fig:dis_hist}. 2) Distribution of agents is uniform w.r.t objects, instead of the field of view (0-280m). Fig.\ref{fig:agent_hist} shows that the distribution of agents is same with all objects.

\begin{figure*}[!t]
\vspace{-7mm}
\centering
\noindent
% \hspace{1mm}
\centering
    \begin{subfigure}{0.32\linewidth}
    \includegraphics[width=0.99\linewidth]{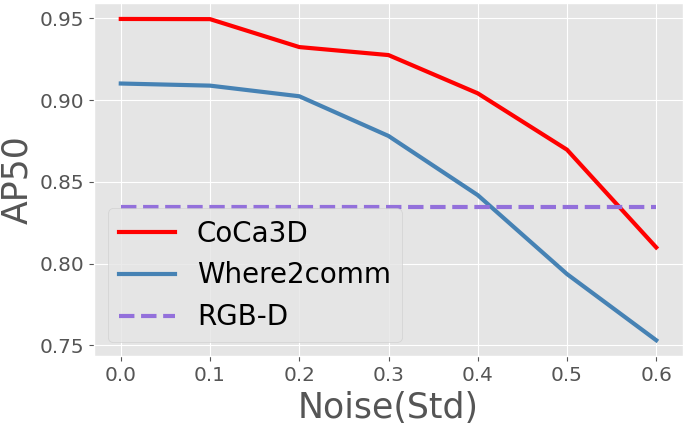}
    % \vspace{-5mm}
    \caption{CoPerception-UAV+}
    \label{fig:noise_uav}
  \end{subfigure}
    \begin{subfigure}{0.32\linewidth}
    \includegraphics[width=0.99\linewidth]{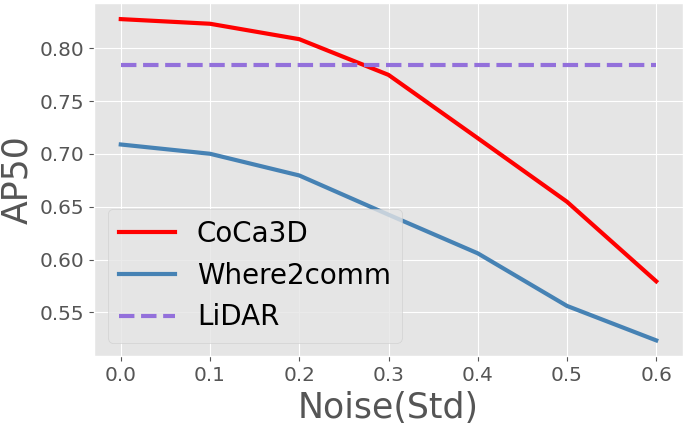}
    % \vspace{-5mm}
    \caption{OPV2V+}
    \label{fig:noise_opv2v}
  \end{subfigure}
  \begin{subfigure}{0.32\linewidth}
    \includegraphics[width=0.99\linewidth]{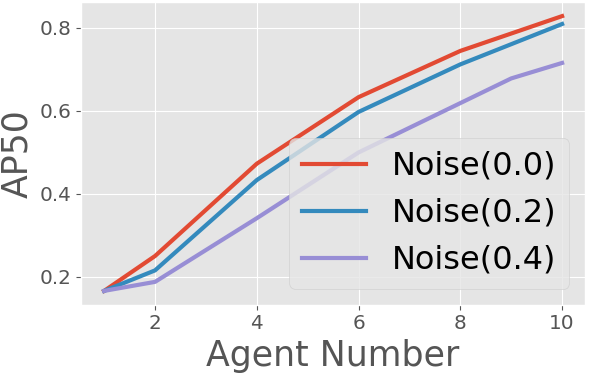}
    % \vspace{-5mm}
    \caption{OPV2V+}
    \label{fig:noise_opv2v_agentnum}
  \end{subfigure}
  % \vspace{-4mm}
  \caption{CoCa3D is as robust as SOTA where2comm (NeurIPS 22) to pose errors.}
% \vspace{-9mm}
\end{figure*}

\subsubsection{CoPerception-UAVs+}
\begin{figure}[!t]
\centering
    \includegraphics[width=0.99\linewidth]{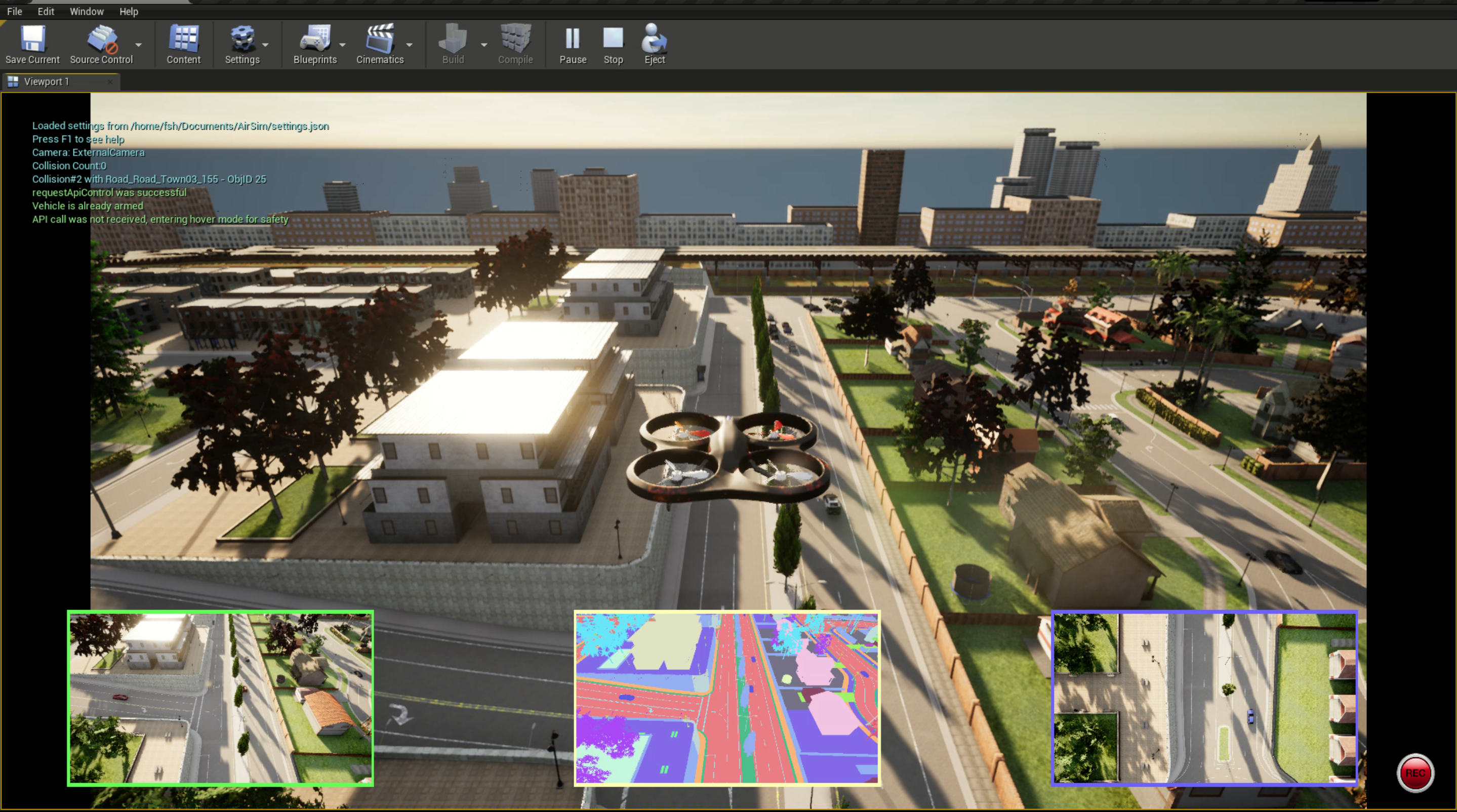}
  \caption{CoPerception-UAVs+ is co-simulated by CARLA~\cite{dosovitskiy2017carla} and AirSim~\cite{Airsim}.}
    \label{fig:uav_env}
\end{figure}

\begin{figure*}[!t]
    \centering
    \begin{subfigure}{0.24\linewidth}
    \includegraphics[width=0.99\linewidth]{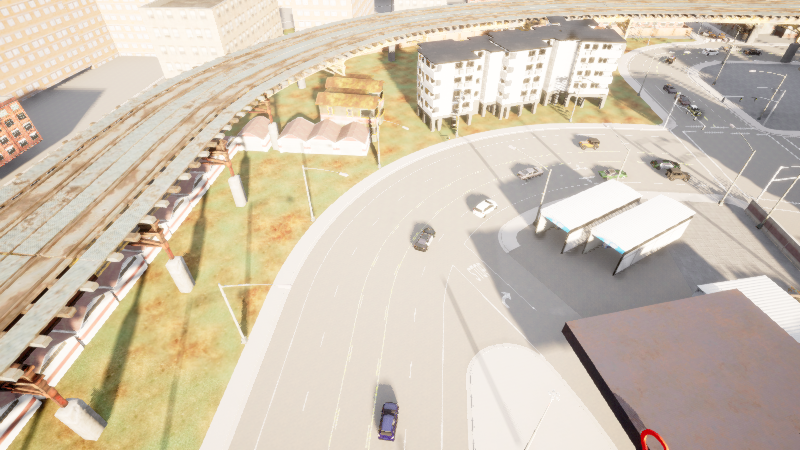}
    \caption{Agent 0: Camera 0}
  \end{subfigure}
  \hfill
  \begin{subfigure}{0.24\linewidth}
    \includegraphics[width=0.99\linewidth]{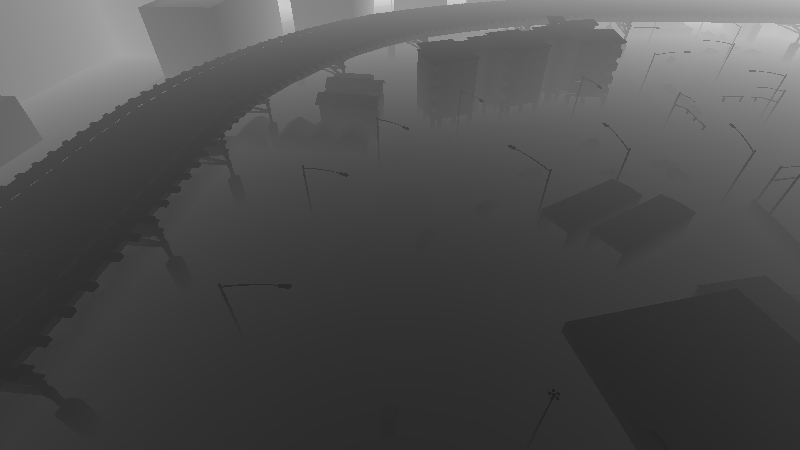}
    \caption{Agent 0: Depth 0}
  \end{subfigure}
  \hfill
  \begin{subfigure}{0.24\linewidth}
    \includegraphics[width=0.99\linewidth]{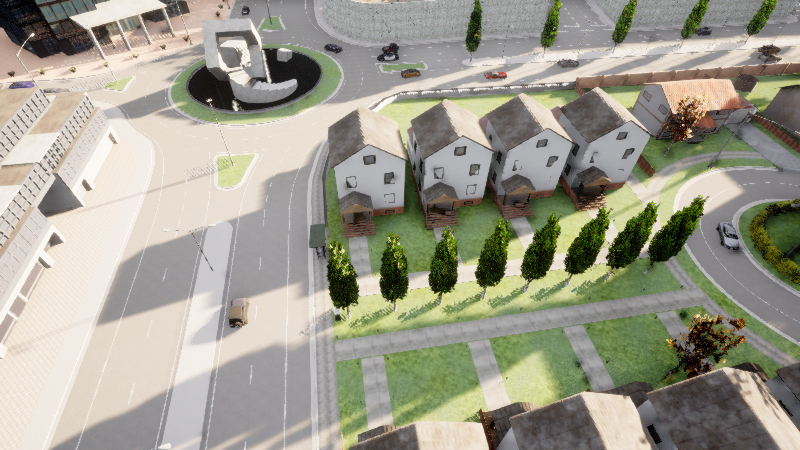}
    \caption{Agent 1: Camera 0}
  \end{subfigure}
  \hfill
  \begin{subfigure}{0.24\linewidth}
    \includegraphics[width=0.99\linewidth]{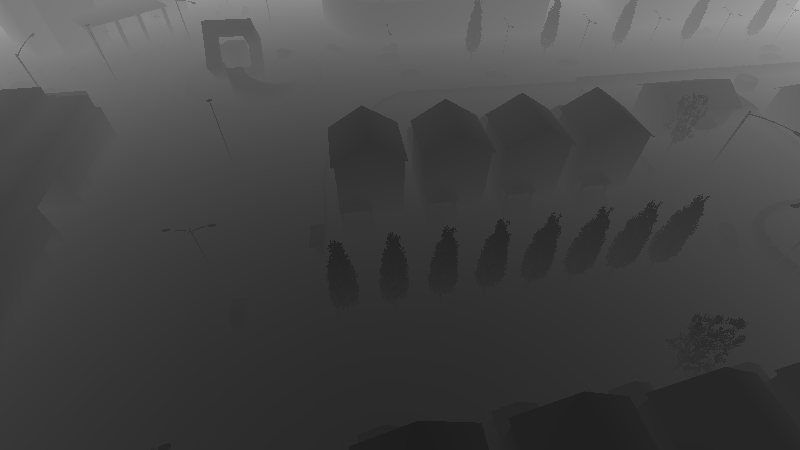}
    \caption{Agent 1: Depth 0}
  \end{subfigure}
  \hfill
  \begin{subfigure}{0.24\linewidth}
    \includegraphics[width=0.99\linewidth]{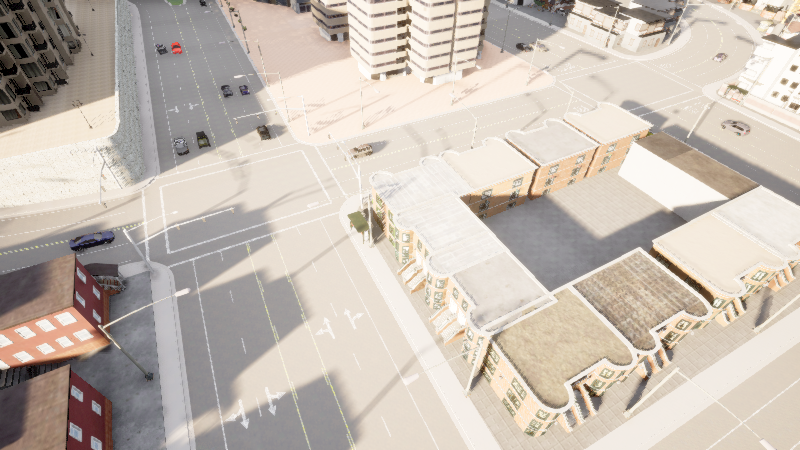}
    \caption{Agent 2: Camera 0}
  \end{subfigure}
  \hfill
  \begin{subfigure}{0.24\linewidth}
    \includegraphics[width=0.99\linewidth]{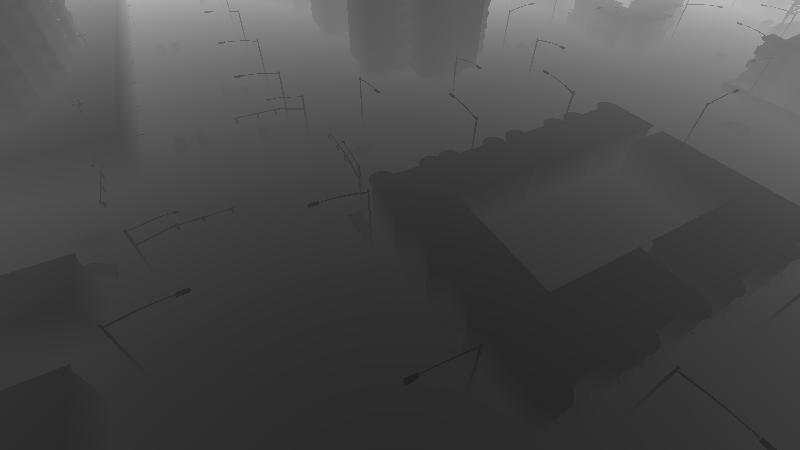}
    \caption{Agent 2: Depth 0}
  \end{subfigure}
  \hfill
  \begin{subfigure}{0.24\linewidth}
    \includegraphics[width=0.99\linewidth]{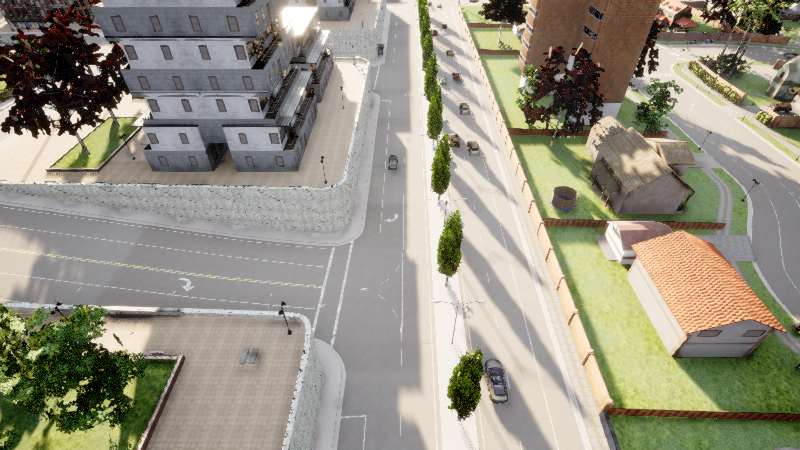}
    \caption{Agent 3: Camera 0}
  \end{subfigure}
  \hfill
  \begin{subfigure}{0.24\linewidth}
    \includegraphics[width=0.99\linewidth]{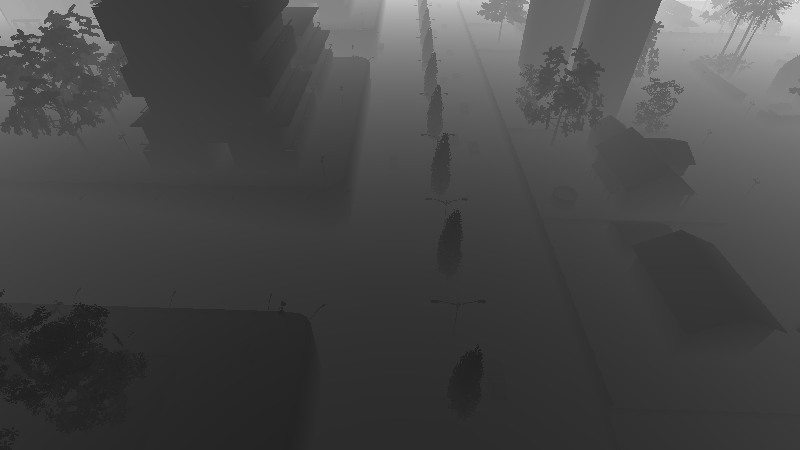}
    \caption{Agent 3: Depth 0}
  \end{subfigure}
  \hfill
  \begin{subfigure}{0.24\linewidth}
    \includegraphics[width=0.99\linewidth]{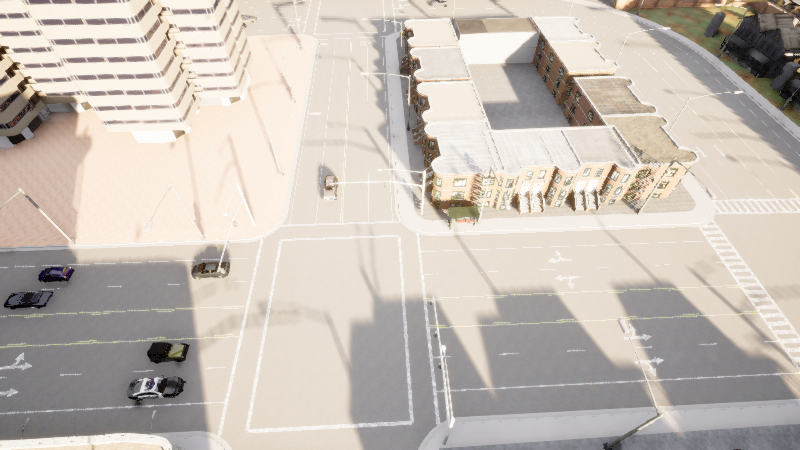}
    \caption{Agent 4: Camera 0}
  \end{subfigure}
  \hfill
  \begin{subfigure}{0.24\linewidth}
    \includegraphics[width=0.99\linewidth]{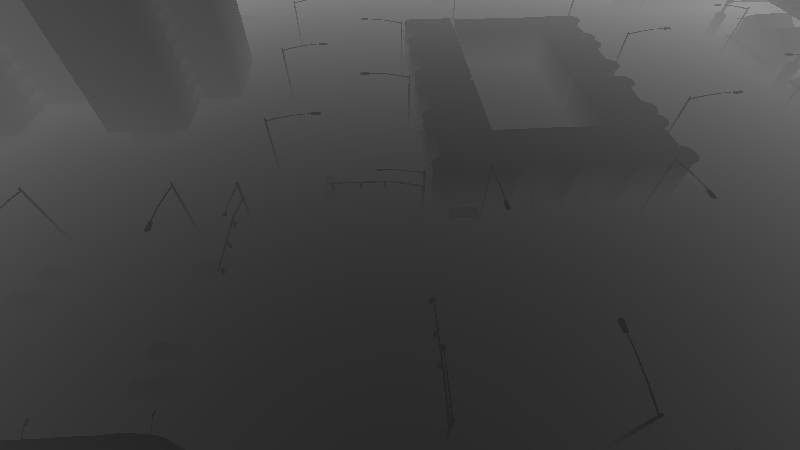}
    \caption{Agent 4: Depth 0}
  \end{subfigure}
  \hfill
  \begin{subfigure}{0.24\linewidth}
    \includegraphics[width=0.99\linewidth]{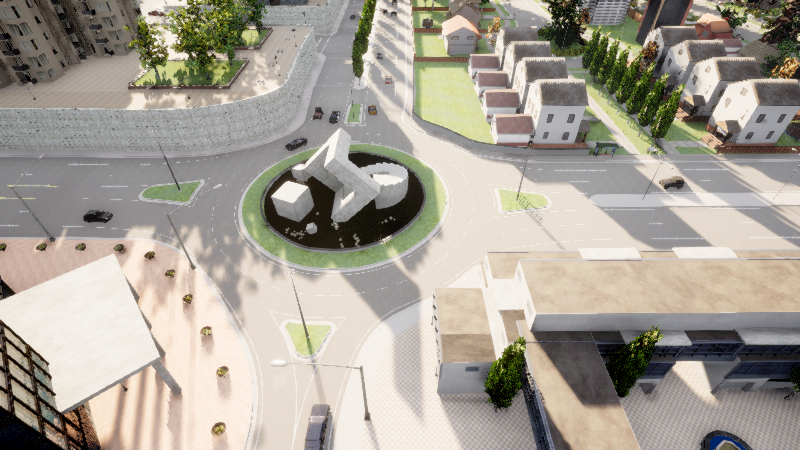}
    \caption{Agent 5: Camera 0}
  \end{subfigure}
  \hfill
  \begin{subfigure}{0.24\linewidth}
    \includegraphics[width=0.99\linewidth]{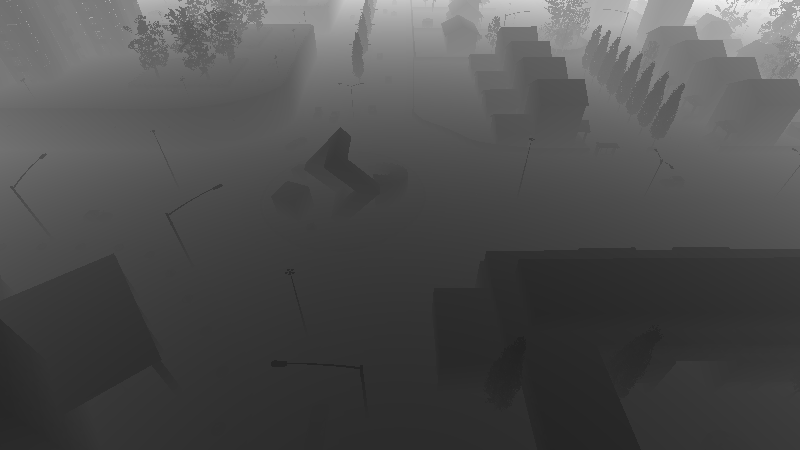}
    \caption{Agent 5: Depth 0}
  \end{subfigure}
  \hfill
  \begin{subfigure}{0.24\linewidth}
    \includegraphics[width=0.99\linewidth]{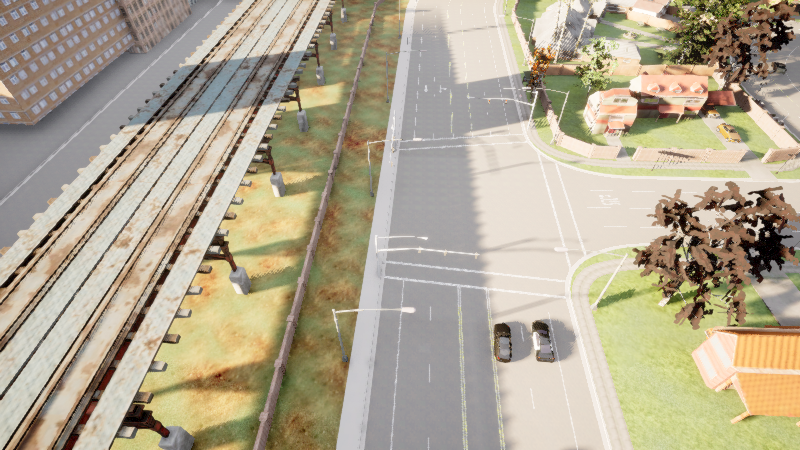}
    \caption{Agent 6: Camera 0}
  \end{subfigure}
  \hfill
  \begin{subfigure}{0.24\linewidth}
    \includegraphics[width=0.99\linewidth]{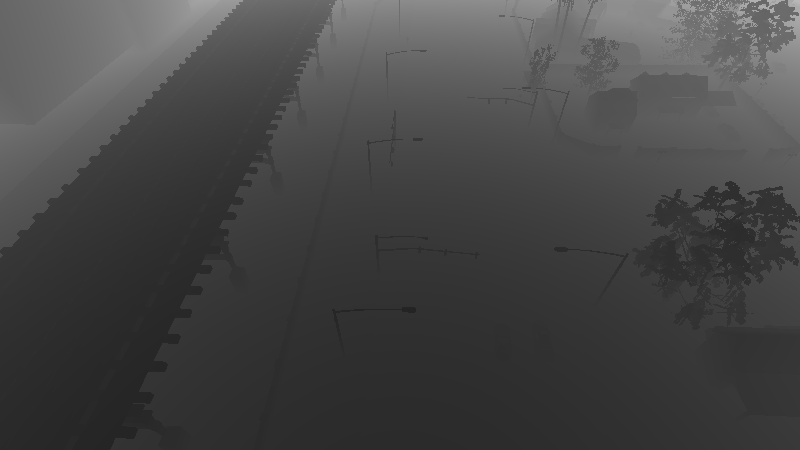}
    \caption{Agent 6: Depth 0}
  \end{subfigure}
  \hfill
  \begin{subfigure}{0.24\linewidth}
    \includegraphics[width=0.99\linewidth]{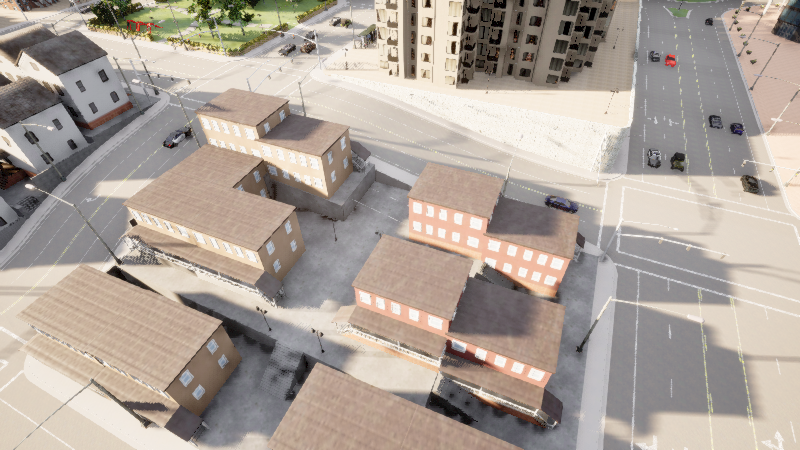}
    \caption{Agent 7: Camera 0}
  \end{subfigure}
  \hfill
  \begin{subfigure}{0.24\linewidth}
    \includegraphics[width=0.99\linewidth]{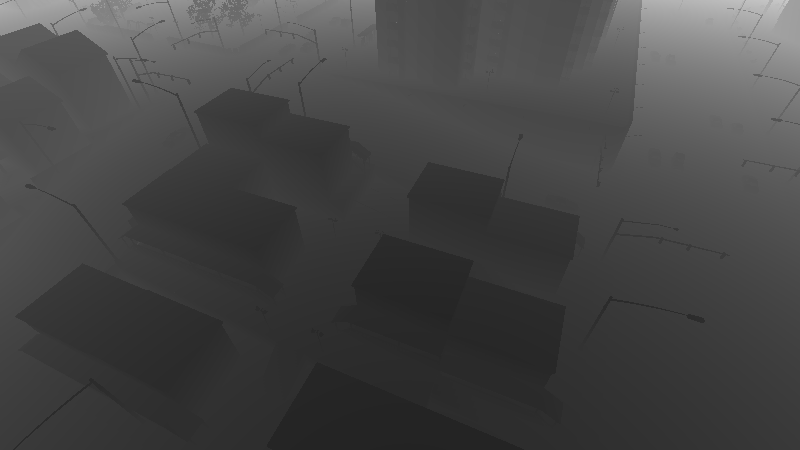}
    \caption{Agent 7: Depth 0}
  \end{subfigure}
  \hfill
  \begin{subfigure}{0.24\linewidth}
    \includegraphics[width=0.99\linewidth]{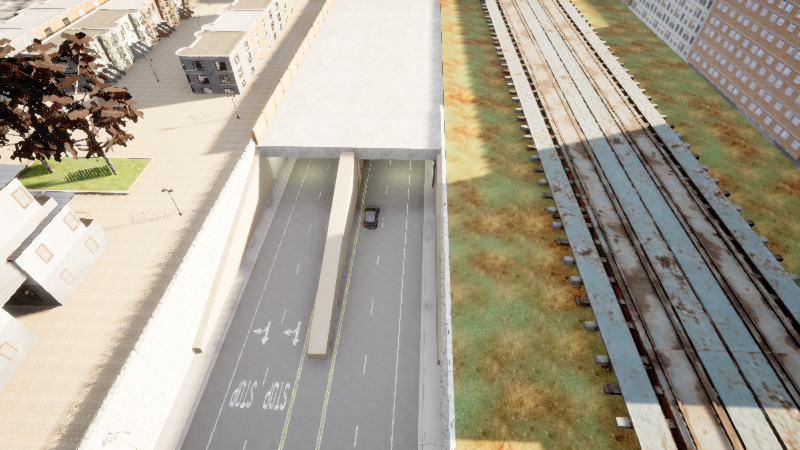}
    \caption{Agent 8: Camera 0}
  \end{subfigure}
  \hfill
  \begin{subfigure}{0.24\linewidth}
    \includegraphics[width=0.99\linewidth]{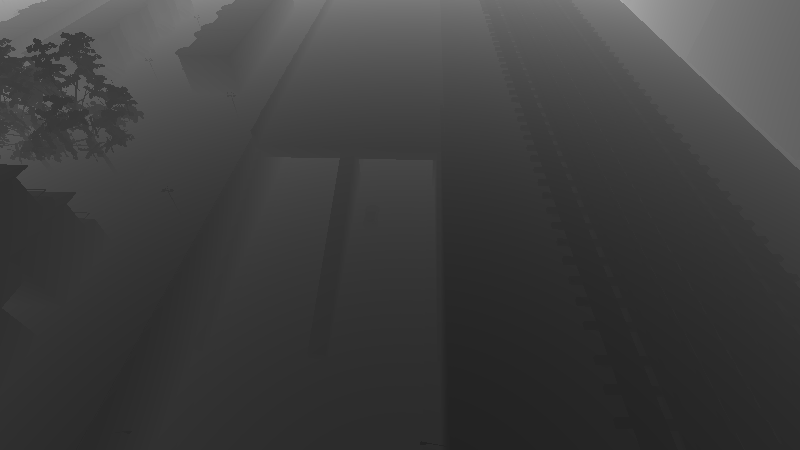}
    \caption{Agent 8: Depth 0}
  \end{subfigure}
  \hfill
  \begin{subfigure}{0.24\linewidth}
    \includegraphics[width=0.99\linewidth]{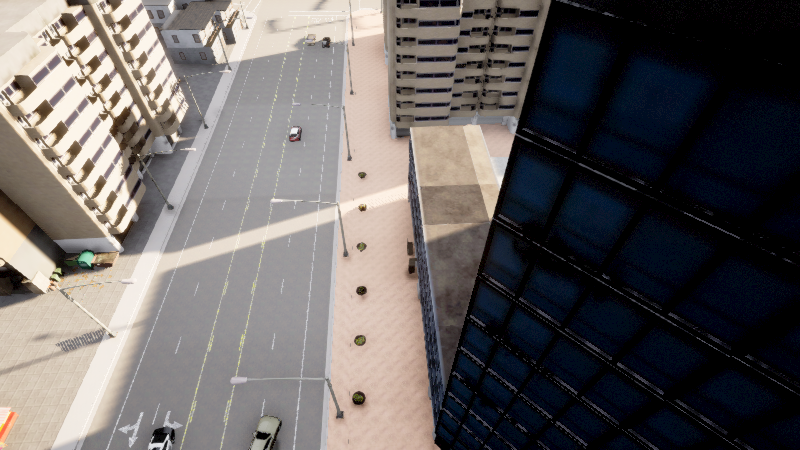}
    \caption{Agent 9: Camera 0}
  \end{subfigure}
  \hfill
  \begin{subfigure}{0.24\linewidth}
    \includegraphics[width=0.99\linewidth]{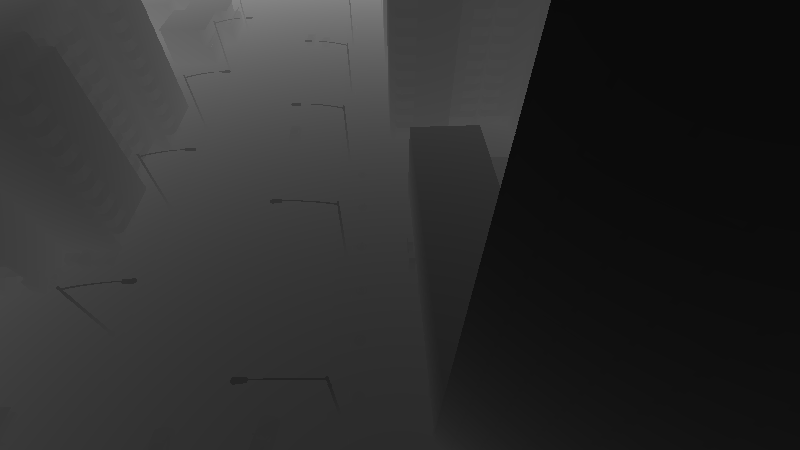}
    \caption{Agent 9: Depth 0}
  \end{subfigure}
  \vspace{-2mm}
  \caption{Data sample with 10 agents of CoPerception-UAVs+.}
  \label{Fig:uav_sample}
  \vspace{-2mm}
\end{figure*}

\noindent
\textbf{Data generation.} We extend the original  CoPerception-UAVs~\cite{hu2022where2comm} with more collaborative agents ($10$). Our CoPerception-UAVs+ is co-simulated by AirSim~\cite{Airsim} and CARLA~\cite{dosovitskiy2017carla}. We use CARLA to generate complex simulation scenes and traffic flow, and AirSim to simulate drones flying in the scene and taking images. And we carefully design the drones' flying route to ensure safety as more agents increase the collision possibility. Figure.~\ref{fig:uav_env} shows the simulation environment. For CoPerception-UAVs+, we simulate more UAVs in AirSim and additionally equip depth sensors for each UAV at the same coordinate with the camera sensor. Figure.~\ref{Fig:uav_sample} shows a randomly selected data sample.

\subsection{Implementation details}
For the camera-only 3D object detection for cars, we implement the detector following the LSS~\cite{philion2020lift} and CaDDN~\cite{CaDDN} for OPV2V+ and DAIR-V2X datasets. We uniformly space the depth into 50 categories. For the training strategy, we first train the single agent detector for 50 epochs with an initial learning rate of $1.5$e-$3$, and decay by 0.1 at epoch 30. Then we load the single pre-trained model and train the whole model with collaboration for another 20 epochs with a learning rate of $1$e-$3$. 

For the camera-only 3D object detection for drones, we implement the detector following the 3D aerial object detection DVDET~\cite{Hu2022AM3D}. We uniformly space the depth into 10 categories. For the training strategy, we train the model for 140 epochs with an initial learning rate of $5$e-$4$, and decay by 0.1 at epoch 80 and 120.

\subsection{Robustness to pose error}
In the paper, we assume collaborative agents' poses are accurate as practical agents should have strong self-localization ability. 
We further assess the model's robustness to agent pose errors. Encouragingly, {\bf CoCa3D still performs well even pose errors appear}. Following the same pose-error setting in Where2comm~\cite{hu2022where2comm} (Gaussian noise with 0 mean and 0m-0.6m standard deviation), our experiments validate CoCa3D's robustness and find: 1) CoCa3D outperforms Where2comm under various pose errors, see Fig.~\ref{fig:noise_uav} and~\ref{fig:noise_opv2v}.
2) CoCa3D still outperforms LiDAR under large pose error 0.5m/0.2m on Co-UAV+/OPV2V+. 3) CoCa3D's performance steadily increases with agent number even with pose errors, see Fig.~\ref{fig:noise_opv2v_agentnum}. Moreover, CoCa3D can integrate customized alignment methods, such as [1,2], to further tackle pose errors.
% , as the pose error is less than 0.2m in most cases
% \textbf{A1}: We indeed assume collaborative agents' poses are accurate as practical agents should have strong self-localization ability. To validate CoCa3D's robustness when pose errors appear, we conduct additional experiments by following the pose-error setting in Where2comm (Gaussian noise with 0 mean and 0m-0.6m standard deviation), encouragingly, we find that {\bf CoCa3D performs equally well}: i) CoCa3D outperforms previous SOTA under various pose-error levels, see Fig.~\ref{fig:noise_uav} and~\ref{fig:noise_opv2v};

% \clearpage